%% file: main.tex
\documentclass[10pt,twocolumn,letterpaper]{article}

 \usepackage[pagenumbers]{cvpr} 

\input{preamble}

\definecolor{cvprblue}{rgb}{0.21,0.49,0.74}
\usepackage[pagebackref,breaklinks,colorlinks,citecolor=cvprblue]{hyperref}
\usepackage[utf8]{inputenc} 
\usepackage[T1]{fontenc}    
\usepackage{hyperref}       
\usepackage{url}            
\usepackage{booktabs}       
\usepackage{amsfonts}       
\usepackage{nicefrac}       
\usepackage{microtype}      
\usepackage{xcolor}         
\usepackage{graphicx}
\usepackage{multirow}
\usepackage{algorithm}
\usepackage{listings}
\usepackage{wrapfig}

\usepackage{arydshln}

\usepackage{rotating}

\def\paperID{6533} 
\def\confName{CVPR}
\def\confYear{2024}

\title{LeftRefill: Filling Right Canvas based on Left Reference through \\
Generalized Text-to-Image Diffusion Model}

\author{Chenjie Cao$^{1,2,3}$\footnotemark[1], Yunuo Cai$^{1}$, Qiaole Dong$^{1}$, Yikai Wang$^{1}$, Yanwei Fu$^{1}$\footnotemark[2]\\
$^1$\normalsize School of Data Science, 
Fudan University\\
$^2$\normalsize Shanghai Key Lab of Intelligent Information Processing, School of Computer Science, Fudan University\\
$^3$\normalsize Alibaba Group\\
{\tt\small \{cjcao20,yncai20,qldong18,yikaiwang19,yanweifu\}@fudan.edu.cn}}

\begin{document}

\twocolumn[{
\renewcommand\twocolumn[1][]{#1}
\maketitle
\begin{center}
\includegraphics[width=0.96\linewidth]{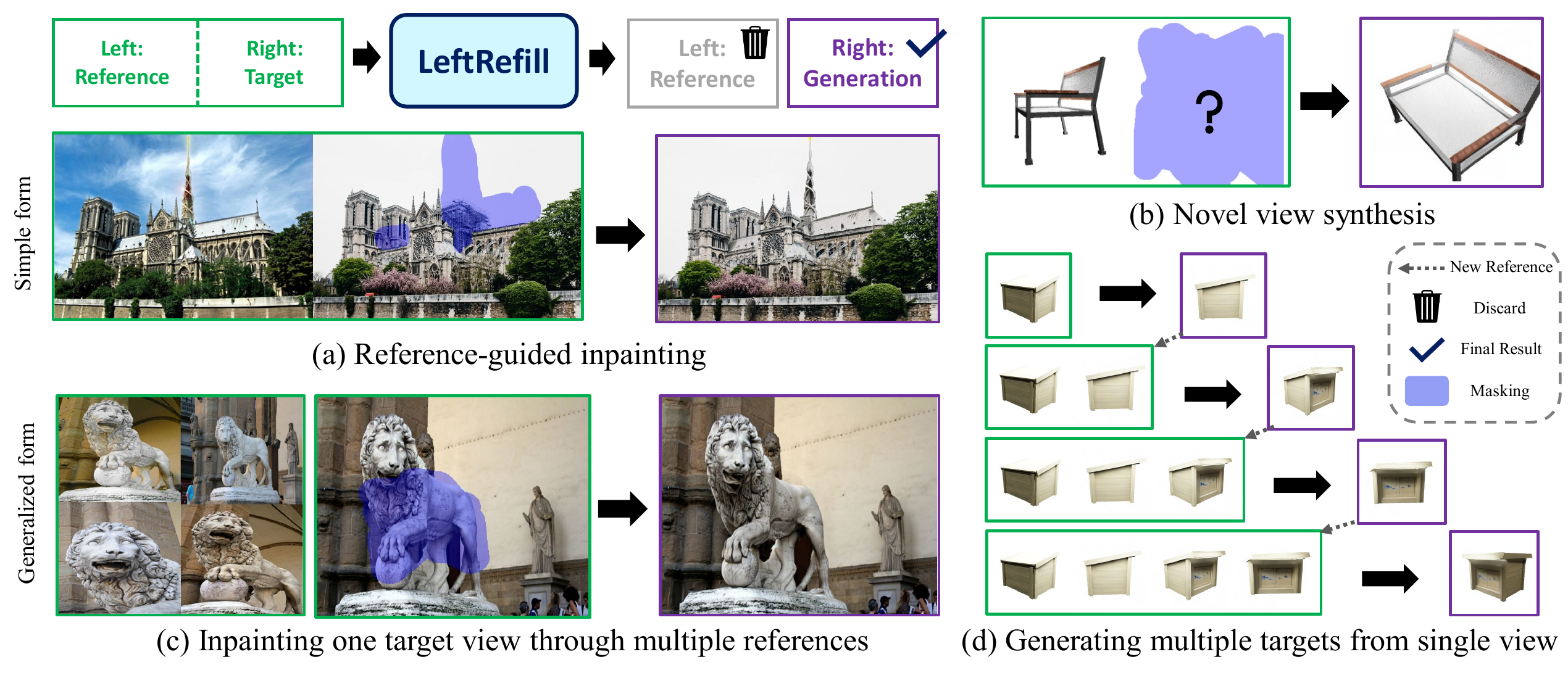}
\vspace{-0.15in}
\captionof{figure}{ 
{LetRefill addresses the generation on the right canvas conditioned by left references.
We can re-formulate several existing tasks in the LeftRefill manner, including (a) reference-guided inpainting, (b) novel view synthesis.
The reference and target can be further extended to multi-view scenes, forming (c) multi-view reference inpainting and (d) multi-view synthesis, respectively.
\textcolor[rgb]{0.0,0.5,0.2}{\textbf{Green}} frames indicate stitched inputs. Reference views are placed on the left side, while masked target views are placed on the right side. \textcolor{violet}{\textbf{Violet}} frames only show enlarged right-side generations produced by LeftRefill.
Note that we omit some input details in (c) and (d) for simplicity. 
}
   \label{fig:teaser}}
\end{center}
}]

\renewcommand{\thefootnote}{\fnsymbol{footnote}}
\footnotetext[1]{This work was accomplished while Cao was at Fudan University.}
\footnotetext[2]{Corresponding author.}

\begin{abstract}
This paper introduces LeftRefill, an innovative approach to efficiently harness large Text-to-Image (T2I) diffusion models for reference-guided image synthesis. 
As the name implies, LeftRefill horizontally stitches reference and target views together as a whole input. 
The reference image occupies the left side, while the target canvas is positioned on the right.
Then, LeftRefill paints the right-side target canvas based on the left-side reference and specific task instructions. 
Such a task formulation shares some similarities with contextual inpainting, akin to the actions of a human painter.
This novel formulation efficiently learns both structural and textured correspondence between reference and target without other image encoders or adapters.
We inject task and view information through cross-attention modules in T2I models, and further exhibit multi-view reference ability via the re-arranged self-attention modules.
These enable LeftRefill to perform consistent generation as a generalized model without requiring test-time fine-tuning or model modifications.
Thus, LeftRefill can be seen as a simple yet unified framework to address reference-guided synthesis. 
As an exemplar, we leverage LeftRefill to address two different challenges: reference-guided inpainting and novel view synthesis, based on the pre-trained StableDiffusion. Codes\&models are released at \url{https://github.com/ewrfcas/LeftRefill}.
\end{abstract}

\section{Introduction}

Imagine you are a right-handed painter with a task that requires you to draw or modify a target image based on a reference picture. How would you approach it? Intuitively, you would likely place the reference image on your left side and paint or modify the target view on the right side conditioned on the left one.
If we regard the large Text-to-Image (T2I) models~\cite{nichol2021glide,rombach2022high,ramesh2022hierarchical,saharia2022photorealistic,chang2023muse,podell2023sdxl} as skillful painters, could they also follow such simple and intuitive task formulation to handle complex reference-guided synthesis tasks?
In this paper, we explore this problem with two challenging tasks: 
1) Inpainting masked target views conditioned on reference images, \ie, reference-guided inpainting (Ref-inpainting)~\citep{zhou2021transfill,zhao2022geofill} as in \Cref{fig:teaser}(a).
2) Generating new views based on known images of specific objects, \ie, Novel View Synthesis (NVS)~\citep{liu2023zero} as in \Cref{fig:teaser}(b).

It seems straightforward to harness the power of T2I generative models to directly address these image-reference tasks by training additional adapters~\citep{hu2021lora,zhang2023adding,mou2023t2i} or replacing textual encoders with visual ones~\cite{yang2023paint,liu2023zero}
and optimize them for full fine-tuning of the entire T2I model.
We should clarify that training these large T2I models with unfamiliar visual encoders is computationally intensive and challenging to converge, particularly when working with limited batch sizes. Additionally, most visual encoders, such as image CLIP~\citep{radford2021learning}, tend to emphasize the learning of semantic features rather than the intricately spatial details that are essential for tasks involving Ref-inpainting as verified in our experiments.

To prevent the heavy modification to T2I models mentioned above, we rethink the human painting habit and propose LeftRefill, a unified approach for both Ref-inpainting and NVS. 
LeftRefill is built upon the inpainting fine-tuned StableDiffusion2.0  (SD)~\citep{rombach2022high}\footnote{\scriptsize\url{https://github.com/Stability-AI/stablediffusion}.}, which ingeniously reformulates reference-based synthesis as a contextual inpainting process.
Specifically, LeftRefill horizontally stitches reference and target views as a whole input. 
As shown in \Cref{fig:teaser}, reference images are positioned on the left side, while masked targets are on the right side. 
This simple yet effective formulation eliminates the dependency on additional image feature encoders, as both reference and target views have been stitched into the same canvas.
As an ``experienced painter'', LeftRefill is driven by specific instructions, called task and view prompt tuning. 
These instructions infuse crucial information for specific generative tasks and desired view orders to cross-attention modules, guiding the generation of SD.
Note that LeftRefill is a generalized framework, so we can train two LeftRefill models for Ref-inpainting and NVS separately without any test-time fine-tuning. 

On the other hand, we emphasize that LeftRefill further enjoys extension into multi-view synthesis scenarios, including image inpainting conditioned on multi-view references in \Cref{fig:teaser}(c) and consistent NVS from a single view in \Cref{fig:teaser}(d).
Formally, we rearrange the tensor shape before the self-attention modules, enabling self-attention to capture cross-view information.
Moreover, to tackle the more intricate NVS task, we propose the novel technique of block causal masking, facilitating self-attention-based T2I models in achieving consistent autoregressive (AR) generation. 
All improvements are integrated into the off-the-shelf SD components, involving no additional image encoder, prompt tuning based on cross-attention, self-attention rearranging and block causal masking. 
LeftRefill emerges as a unified architecture for addressing reference-guided synthesis, requiring only minor model modifications.

We highlight the key contributions of LeftRefill as follows:
1) \emph{Lightweight and generalized task formulation based on off-the-shelf T2I models:}
Benefiting from the novel contextual inpainting formulation and inherent attention mechanisms from SD, LeftRefill provides an efficient solution for reference-guided synthesis without thoroughly re-training T2I models and test-time fine-tuning. 
2) \emph{Task and view-specific prompt tuning:} 
Our work pioneers the use of task and view-specific prompt tuning, allowing for precise control over generative tasks and view orders.
3) \emph{End-to-end Ref-inpainting:}
Notably, our LeftRefill addresses the challenging Ref-inpainting end-to-end, without complex 3D geometrical warping and 2D inpainting techniques~\citep{zhou2021transfill,zhao2022geofill,zhao20223dfill}. 
4) \emph{Autoregressive NVS with block causal masking}: For the intricate NVS task, we introduce the novel concept of block causal masking, enabling self-attention-based T2I models to achieve AR generation for geometric consistency.


\section{Related Work}

\noindent\textbf{Personalization and Controllability of T2I Models.}
Recent achievements on T2I have produced impressive visual generations~\citep{mou2023t2i,bar2023multidiffusion,poole2023dreamfusion}, which could be further extended into local editing~\citep{avrahami2022blended,hertz2022prompt,couairon2022diffedit}.
However, these models could only be controlled by natural languages. As ``an image is worth hundreds of words'', T2I models based on natural texts fail to produce images with specific textures, locations, identities, and appearances~\citep{gal2022image}. Textual inversion~\citep{gal2022image,mokady2022null} and fine-tuning techniques~\citep{ruiz2022dreambooth} are proposed for personalized T2I.
Meanwhile, many works pay attention to image-guided generations~\citep{voynov2022sketch,li2023gligen,ma2023unified}. ControlNet~\citep{zhang2023adding} and T2I-Adapter~\citep{mou2023t2i} learn trainable adapters~\citep{houlsby2019parameter} to inject visual clues to pre-trained T2I models without losing generalization and diversity.
But these moderate methods only work for simple style transfers. More spatially complex tasks, such as Ref-inpainting, are difficult to handle by ControlNet as verified in \Cref{sec:experiments}.
In contrast, T2I-based exemplar-editing and NVS have to be fine-tuned on large-scale datasets with strong data augmentation~\citep{yang2023paint} and large batch size~\citep{liu2023zero}.
Compared with these aforementioned manners, LeftRefill enjoys both spatial modeling capability and computational efficiency.

\noindent\textbf{Prompt Tuning} ~\citep{lester2021power,liu2021gpt,liu2021p} indicates fine-tuning token embeddings for transformers with frozen backbone to preserve the capacity. Prompt tuning is first explored for adaptively learning suitable prompt features for language models rather than manually selecting them for different downstream tasks~\citep{liu2023pre}. 
Moreover, prompt tuning has been further investigated in vision-language models~\citep{radford2021learning,ge2022domain} and discriminative vision models~\citep{jia2022visual,liao2023rethinking}.
Visual prompt tuning in~\cite{sohn2022visual} prepends trainable tokens before the visual sequence for transferred generations. 
Though both LeftRefill and~\cite{sohn2022visual} aim to tackle image synthesis, our prompt tuning is used for controlling text encoders rather than visual ones. 
Thus LeftRefill enjoys more intuitive prompt initialization from task-related textual descriptions.

\noindent\textbf{Reference-guided Image Inpainting.}
Image inpainting is a long-standing vision task, which aims to fill missing image regions with coherent results. Both traditional methods~\citep{bertalmio2000image,criminisi2003object,hays2007scene} and learning-based ones~\citep{zeng2020high,zhao2021large,li2022mat,suvorov2022resolution,dong2022incremental} achieved great progress in image inpainting. 
Furthermore, Ref-inpainting requires recovering a target image with one or several reference views from different viewpoints~\citep{oh2019onion}, which is useful for repairing old buildings or removing occlusions in popular attractions.
But Ref-inpainting usually suffers from a sophisticated pipeline~\citep{zhou2021transfill,zhao2022geofill,zhao20223dfill}, including depth estimation, pose estimation, homography warping, and single-view inpainting. 
Limited by large holes, the estimated geometric pose is not reliable, largely degrading the pipeline. Thus an end-to-end Ref-inpainting pipeline is meaningful.
To the best of our knowledge, we are the first ones to tackle such a difficult reference-guided task with T2I models.

\noindent\textbf{Novel View Synthesis from a Single Image.}
NVS based on a single image is an intractable ill-posed problem, requiring both sufficient geometry understanding and expressive textural presentation~\citep{fahim2021single}.
Many previous works could partially tackle this problem through single view 3D reconstruction~\citep{wu2017marrnet,wang2018pixel2mesh,chen2019learning,liu2019soft,xu2019disn}, 2D generative models~\citep{niklaus20193d,shih20203d,rombach2021geometry}, feature cost volumes~\citep{chan2023generative}, and GAN-based methods~\citep{schwarz2020graf,niemeyer2021giraffe,chan2022efficient}.
However, these manners still suffer from limited generalization or small angle variations. 
To address this issue, Zero123~\citep{liu2023zero} uses another image CLIP encoder to inject image features from the reference view to unlock the capacity from 2D diffusion-based T2I models for NVS. But Zero123 requires a large batch size and expensive computational resources to stabilize the training stage with an unknown reference image encoder. Moreover, the image encoder in Zero123 can only tackle one reference image, which fails to generate consistent multi-view images.

\section{Method}

\begin{figure}
\centering
\includegraphics[width=1.0\linewidth]{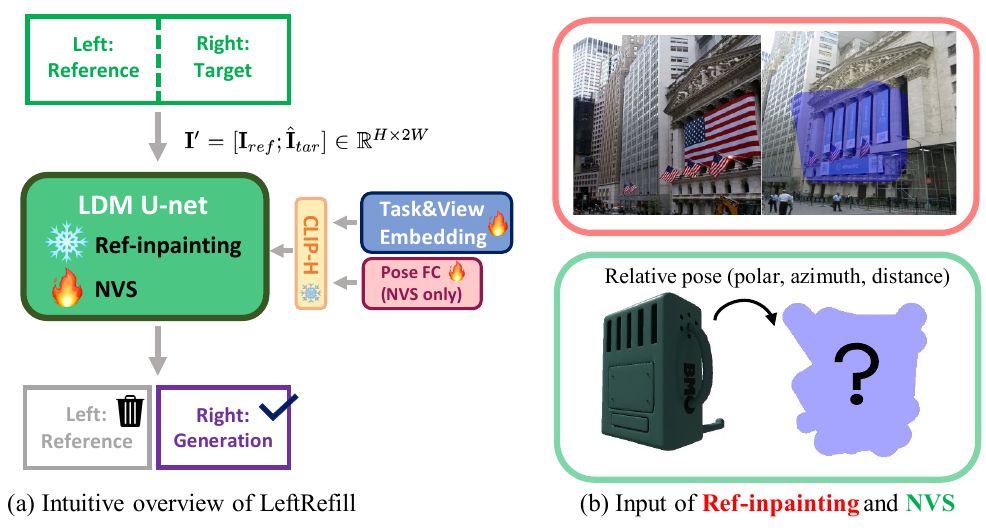}
\vspace{-0.2in}
   \caption{(a) The overview of LeftRefill.
   Inputs of Ref-inpainting and NVS are shown in (b). Task and view prompt embedding and pose features (optional for NVS) are infused to CLIP-H for cross-attention learning in U-net.
   For the output of LeftRefill, we discard the left-side reference and take the right-side generation.
   \label{fig:model_overview}}
\vspace{-0.15in}
\end{figure}

\begin{figure}
\centering
\includegraphics[width=0.9\linewidth]{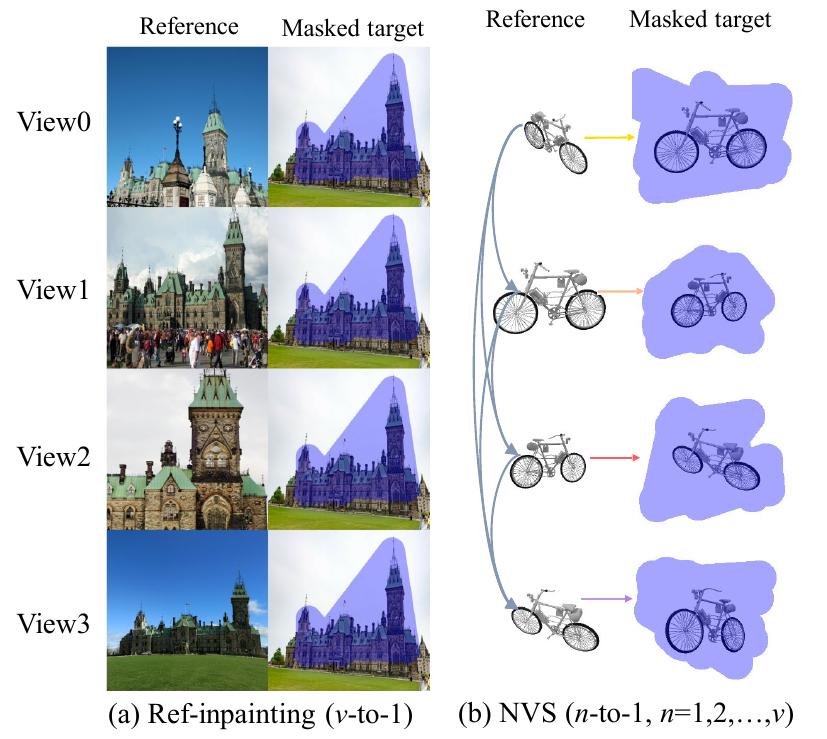}
\vspace{-0.15in}
   \caption{  Illustration about multi-view training inputs ($v\times H\times 2W$, $v=4$) of LeftRefill, where $v,H,2W$ indicate the view number, height, and width of stitching images. 
   All views of Ref-inpainting (a) share the same masked target, while the multi-view NVS (b) should be trained with the AR generation. 
\label{fig:task_formulation_and_masking_a}}
\vspace{-0.25in}
\end{figure}

\textbf{Overview.} 
LeftRefill is first formulated in \Cref{sec:framework}. 
Then we explain using self-attention to capture multi-view correspondence, and AR generation based on block casual masking (\Cref{sec:reactivate_self_attn}).
Finally, we discuss the task and view-specific prompt tuning for cross-attention (\Cref{sec:prompt_tuning}).

\subsection{Framework of LeftRefill}
\label{sec:framework}

\begin{figure}
\centering
\includegraphics[width=0.8\linewidth]{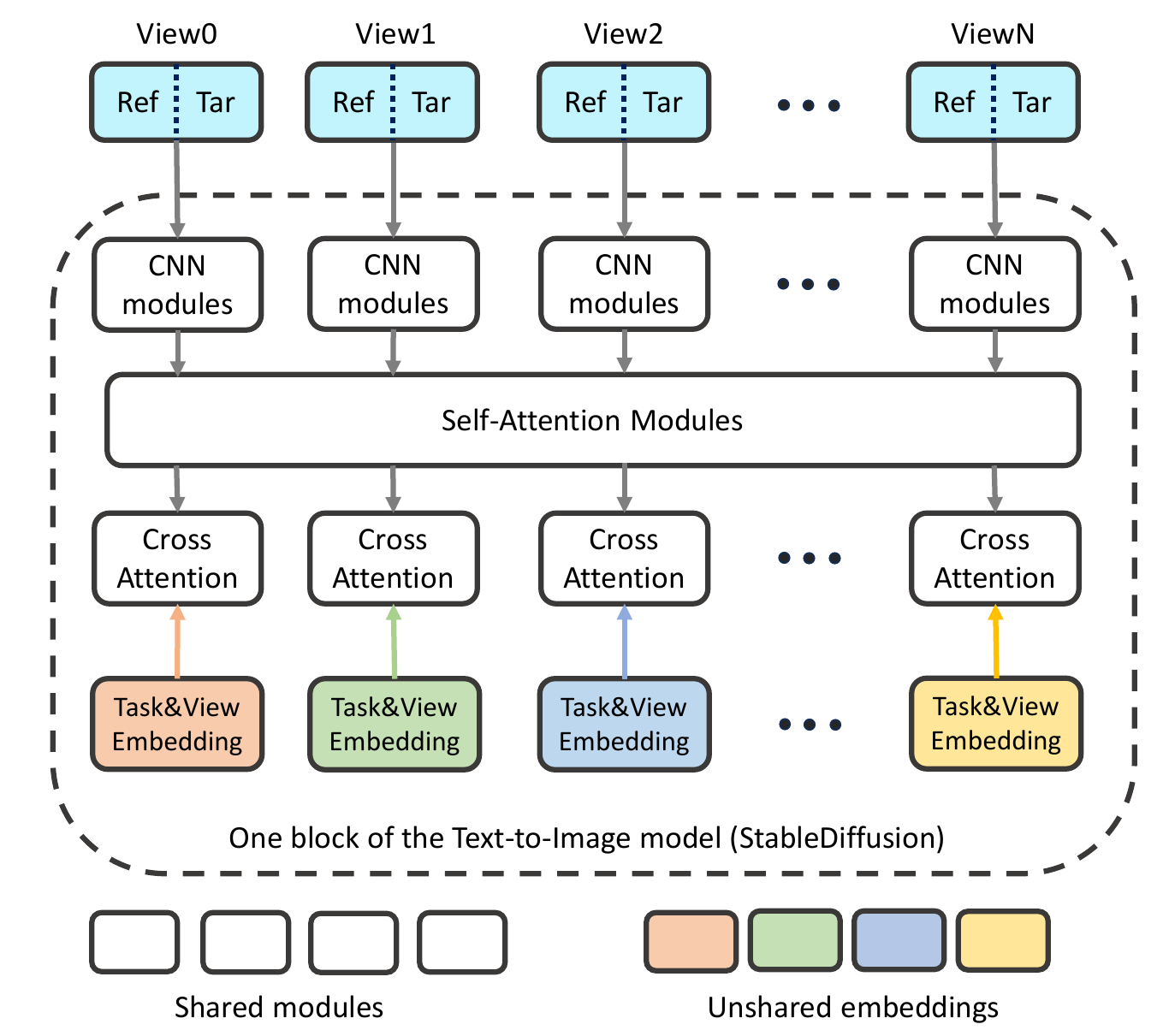}
\vspace{-0.15in}
   \caption{
   Detailed architecture of LeftRefill for multi-view synthesis. Both CNN and cross-attention modules are encoded separately for each stitched view, while all views share the same self-attention for multi-view correlation learning.
   \label{fig:detailed_pipeline}}
\vspace{-0.15in}
\end{figure}

\begin{figure}
\centering
\includegraphics[width=0.85\linewidth]{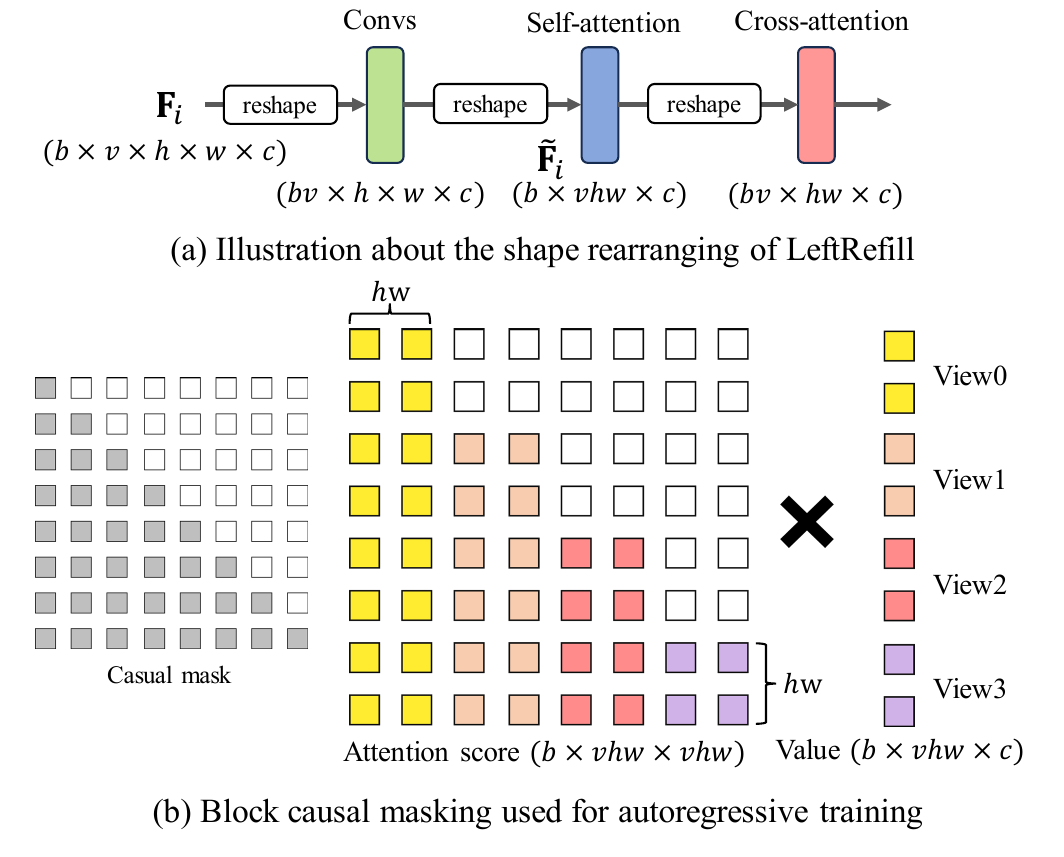}
\vspace{-0.15in}
   \caption{ (a) Feature rearranging, and (b) block causal masking of LeftRefill, where $b,v,h,w,c$ indicate the batch size, view number, height, width, and channels of features, where $w$ is the width of stitching features (downsampled from $2W$).
\label{fig:task_formulation_and_masking_b}}
\vspace{-0.25in}
\end{figure}

\noindent\textbf{Motivations.}
As depicted in \Cref{fig:model_overview}(a),  our LeftRefill is built upon the inpainting fine-tuned SD~\citep{rombach2022high}.
There are two primary motivations that make us stitch reference and target images together and reformulate both Ref-inpainting and NVS as a contextual image inpainting problem.
1) LeftRefill just considers a single input image, eliminating the requirement of additional image encoders and avoiding major architectural alterations and extensive re-training.
2) Since all T2I models are only pre-trained on single-view images, the left-right stitched input could implicitly reactivate the \emph{essential capacity from large T2I models of modeling the correlation of the single-view image}.
Particularly, the left-right stitching input enables the self-attention modules to focus on correct parts from left-side reference at the early stage of diffusion sampling (\Cref{fig:attn_analysis}).
We also thoroughly evaluate different alterations of reference-guided SD in \Cref{sec:exp_ref_inpainting}. 
LeftRefill substantially outperforms other competitors with high efficiency and negligible trainable weights. 


\noindent\textbf{Single-view Formulation (1-to-1).}
Thanks to the convolutional U-net architecture in the Latent Diffusion Model (LDM), we can expand the input image in the spatial dimensions without any modification.
Let's consider a scenario with a single reference image at first, \ie, 1-to-1. Our input $\mathbf{I}'$ is a stitching image of $\mathbf{I}_{ref}$ and the masked target $\hat{\mathbf{I}}_{tar}$, forming as $\mathbf{I}'=[\mathbf{I}_{ref};\hat{\mathbf{I}}_{tar}]\in\mathbb{R}^{H\times 2W}$ as shown in \Cref{fig:model_overview}(a).
In practice, we take the reference image on the left side, while the target one is placed on the right side.
We just take the right-side output as the final generation while the left-side output is discarded.
Note that the diffusion optimization is based on the whole stitched image without any modification.
For the masked target $\hat{\mathbf{I}}_{tar}$, we treat Ref-inpainting's input targets as \emph{partially masked} images, while NVS's input targets are \emph{entirely masked} through the objective segmentation and bounding box.
More details about the processing of data and masks are discussed in the supplementary.

\noindent\textbf{Multi-view Formulation (V-to-1).}
For the multi-view references, we stitch each reference with the specific target as shown in \Cref{fig:task_formulation_and_masking_a}. 
All views are learned separately for convolutions and cross-attention, while they share the same self-attention processing as shown in \Cref{fig:detailed_pipeline} and detailed in \Cref{sec:reactivate_self_attn}.
From \Cref{fig:task_formulation_and_masking_a}(a), the multi-view Ref-inpainting leverages information from different reference views to repair the same target, \ie, $v$-to-1, where $v$ means the view number.
We simply take the generation of the first view as the final inpainted output in $v$-to-1.
For the multi-view NVS, it could be seen as an AR generation for sequentially consistent view synthesis, as depicted in \Cref{fig:task_formulation_and_masking_a}(b), \ie, $n$-to-1 ($n$=1,2,...,$v$).
During the inference, we apply the generated targets as new references for the subsequent view synthesis, while the training phase is accomplished parallelly as detailed in the block casual masking of \Cref{sec:reactivate_self_attn}.

\noindent\textbf{Controlling Generation for LeftRefill.}
Although the self-attention modules in SD have the potential to enable the correlation between left reference and right target, SD is not trained to explicitly activate this capacity. 
The most intuitive way to guide the SD in capturing the correlation among left-right stitched images is to apply suitable text prompts to drive the diffusion model for the desired generation.
However, it is non-trivial to define Ref-inpainting and NVS with natural languages.
Furthermore, it is beneficial to have a non-instance-level prompt to generally guide the diffusion model to accomplish specific tasks.
To this end, we propose to use prompt tuning to learn task and view-specific prompts as detailed in \Cref{sec:prompt_tuning}.
Except for the prompt tuning, all weights in LDM are frozen in Ref-inpainting to maintain the proper generalization as shown in \Cref{fig:model_overview}(a). 
For NVS, we fine-tune the whole LDM to achieve essentially precise pose control, but LeftRefill enjoys much better convergence compared with other fine-tuning based methods~\citep{liu2023zero}. 

\subsection{Reactivating Self-Attention for Multi-View}
\label{sec:reactivate_self_attn}
As shown in \Cref{fig:task_formulation_and_masking_b}(a), given multi-view features $\mathbf{F}_i$ of layer $i$, all MLP, convolutional, and cross-attention layers encode $\mathbf{F}_i$ separately. 
We can simply achieve the separate feature encoding by reshaping the view $v$ and batch $b$ dimension together as $bv$.
Before the self-attention, we rearrange the feature shape as $\mathbf{\tilde{F}}_i\in\mathbb{R}^{b\times vhw\times c}$, thus features across different views could be learned together.
To further incorporate positional clues for distinguishing different sides of reference and target in NVS, we incrementally add positional encoding $P_{i}$ to $\mathbf{\tilde{F}}_i$ before each self-attention block as
\begin{equation}
P_{i}=\gamma_i\cdot\mathrm{cat}([P_{v};P_{Fourier}]),
\label{eq:PE}
\end{equation}
where $P_{v},P_{Fourier}$ indicate the trainable view embedding and Fourier absolute positional encoding~\citep{vaswani2017attention} respectively; $\gamma_i$ is a zero-initialized learnable coefficient for each layer.

For the Ref-inpainting, no masking strategy should be considered in self-attention modules. All reference views share the same target one, thus it is unnecessary for LeftRefill to sequentially repair target views. 
In contrast, for the multi-view NVS, generating consistent novel views from a single image needs our model to handle the sequential generation with dynamic reference views.
For example, the same LeftRefill should accomplish the NVS from one view, two views, and even more. So the AR generation~\citep{van2016conditional,salimans2017pixelcnn++,esser2021taming} is suitable to formulate this task.

\noindent\textbf{Block Causal Masking.}
LeftRefill requires a certain fine-tuning for LDM to effectively tackle challenging NVS as shown in \Cref{fig:model_overview}(a). 
The intuitive solution is to train an AR-based generative model that can generalize across various view numbers for multi-view synthesis.
Converting a pre-trained diffusion model to an AR-based generative model is non-trivial. 
However, the inpainting formulation of LeftRefill makes this conversion feasible. 
Specifically, we just need to adjust the masking strategy during the self-attention learning.
We propose the block casual masking as shown in \Cref{fig:task_formulation_and_masking_b}(b), while the block side-length of each view is $hw$, matching the size of the stitched reference and target pair. Different from the traditional casual mask which is a lower triangular matrix, the block casual mask enlarges the minimal unit from one token to a $hw\times hw$ block, ensuring reasonable block-wise receptive fields. 
In practice, all uncolored tokens in the attention score are masked with ``$-\inf$'' before the softmax operation.
The block casual mask can be implemented parallelly and efficiently as in supplementary. 

\subsection{Task\&View Prompt Tuning} 
\label{sec:prompt_tuning}
The prompt embedding is adopted as the textual branch to CLIP-H~\cite{radford2021learning} of SD, being applied to cross-attention as shown in \Cref{fig:model_overview}(a). 
Specifically, we prepare a set of trainable text embeddings for different generative tasks, which are further categorized into task and view prompts.
Specifically, task prompt embeddings are shared in the same task, \eg, all views of Ref-inpainting using the same task embeddings. In contrast, different view prompt embeddings are applied to inject various view-order information through cross-attention modules to the specific input view.
Though there are only a few trainable parameters (0.05M to 0.065M), we astonishingly find that prompt tuning is sufficient to drive complex generative tasks such as Ref-inpainting, even with a frozen LDM backbone. 
The trainable task and view prompt embeddings $p_t, p_v$ are initialized as the averaged embedding of the natural task description. The optimization target is:
\begin{equation}\footnotesize
\negthickspace\{p_t,p_v\}_{*}=\mathop{\arg\min}\limits_{\{p_t, p_v\}}\mathbb{E}\left[\left\|\varepsilon-\varepsilon_\theta\left([z_t;\hat{z}_0;\mathbf{M}],c_{\phi}(p_t,p_v),t\right)\right\|^2\right],
\label{eq:pt_objective}
\end{equation}
where $\varepsilon_{\theta}(\cdot)$ is the estimated noise by LDM; $c_{\phi}(\cdot)$ means the frozen CLIP-H; $z_t$ is a noisy latent feature of input $z_0$ in step $t$; $\hat{z}_0=z_0\odot(1-\mathbf{M})$ are masked latent features concatenated to $z_t$ with mask $\mathbf{M}$.
Task and view-specific prompt tuning enjoy not only training efficiency but also lightweight saving~\citep{lester2021power}. 
For example, we share the same LeftRefill to address Ref-inpainting with different reference views, while only 0.01\% additional weights of $\{p_t,p_v\}_{*}$ are changed for each view condition.
In NVS, we further provide relative poses to LeftRefill. Following~\cite{liu2023zero}, we calculate the 4-channel relative pose in the polar coordinate for each view, which is encoded by a two-layer FC. Then the pose feature replaces the last padding token in the prompt embeddings before being applied to the CLIP-H.

\section{Experiments}
\label{sec:experiments}

\begin{table} \small 
\vspace{-0.15in}
\caption{Quantitative results for Ref-inpainting on MegaDepth~\citep{li2018megadepth} test set based on matching and manual masks (upper: 1-view; lower: multi-view). 
`ExParams': the scale of extra trainable parameters. * means that the uncorrupted ground truth is visible for the matching. `No stitching': reference and target views are separate without spatial stitching, and only self-attentions are learned across them.
\label{tab:reference_inpainting_quantitative}}
\vspace{-0.125in}
\centering
\footnotesize
\renewcommand\tabcolsep{2pt}
\begin{tabular}{lccccl}
\toprule
Methods & PSNR$\uparrow$ & SSIM$\uparrow$ & FID$\downarrow$ & LPIPS$\downarrow$ & ExParams\tabularnewline
\midrule
SD (inpainting)~\citep{rombach2022high} & 19.841 & 0.819 & 30.260 & 0.1349 & +0\tabularnewline
ControlNet~\citep{zhang2023adding} & 19.072 & 0.744 & 33.664 & 0.1816 & +364.2M\tabularnewline
ControlNet+NewCrossAttn & 19.027 & 0.743 & 34.170 & 0.1805 & +463.4M\tabularnewline
ControlNet+Matching{*}~\citep{tang2022quadtree} & 20.592 & 0.763 & 29.556 & 0.1565 & +364.3M\tabularnewline
Perceiver+ImageCLIP~\citep{jaegle2021perceiver} & 19.338 & 0.745 & 32.911 & 0.1751 & +52.01M\tabularnewline
Paint-by-Example~\citep{yang2023paint} & 18.351 & 0.797 & 34.711 & 0.1604 & +865.9M\tabularnewline
TransFill~\citep{zhou2021transfill} & \textbf{22.744} & \textbf{0.875} & 26.291 & 0.1102 & --\tabularnewline
\hdashline
LeftRefill (no stitching) & 20.489 & 0.827 & \underline{20.125} & \underline{0.1085} & +0.05M\tabularnewline
LeftRefill & \underline{20.926} & \underline{0.836} & \textbf{18.680} & \textbf{0.0961} & +0.05M\tabularnewline
\midrule 
LeftRefill (2-view) & 21.092 & 0.836 & 18.389 & 0.0969 & +0.055M\tabularnewline
LeftRefill (3-view) & 21.356 & 0.840 & 16.838 & 0.0901 & +0.06M\tabularnewline
LeftRefill (4-view) & \textbf{21.779} & \textbf{0.847} & \textbf{16.632} & \textbf{0.0839} & +0.065M\tabularnewline
\bottomrule 
\end{tabular}
\vspace{-0.25in}
\end{table}

\noindent\textbf{Datasets.}
For Ref-inpainting, we use the resized 512$\times$512 image pairs from MegaDepth~\citep{li2018megadepth}, which includes many multi-view famous scenes collected from the Internet. To trade-off between the image correlation and the inpainting difficulty, we empirically retain image pairs with 40\% to 70\% co-occurrence with about 80k images and 820k pairs. 
The validation of Ref-inpainting also includes some manual masks from ETH3D scenes~\citep{schops2017multi} to verify the generalization. 
For the NVS, we use Objaverse~\citep{deitke2022objaverse} rendered by~\cite{liu2023zero} including 800k various scenes with object masks. We resize all images to 256$\times$256 as~\cite{liu2023zero}. Note that some extreme views with elevation angles less than -10$^{\circ}$ are filtered due to excessive ambiguity.
More details about the masking and datasets are introduced in the supplementary.

\begin{figure*}
\begin{center}
\includegraphics[width=0.875\linewidth]{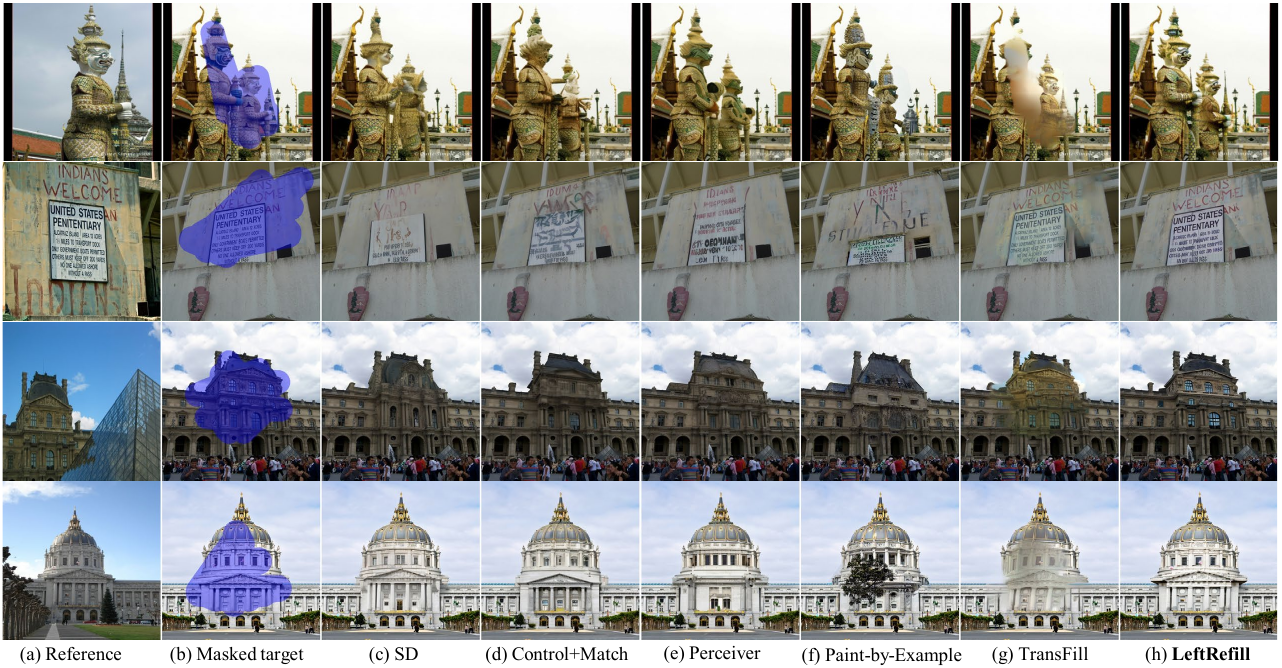}
\vspace{-0.15in}
\end{center}
   \caption{ Qualitative Ref-inpainting results on MegaDepth~\cite{li2018megadepth}.
   More results are in the supplementary.
   \label{fig:reference_guided_qualitative}}
\vspace{-0.15in}
\end{figure*}

\begin{figure}
\begin{center}
\includegraphics[width=1.0\linewidth]{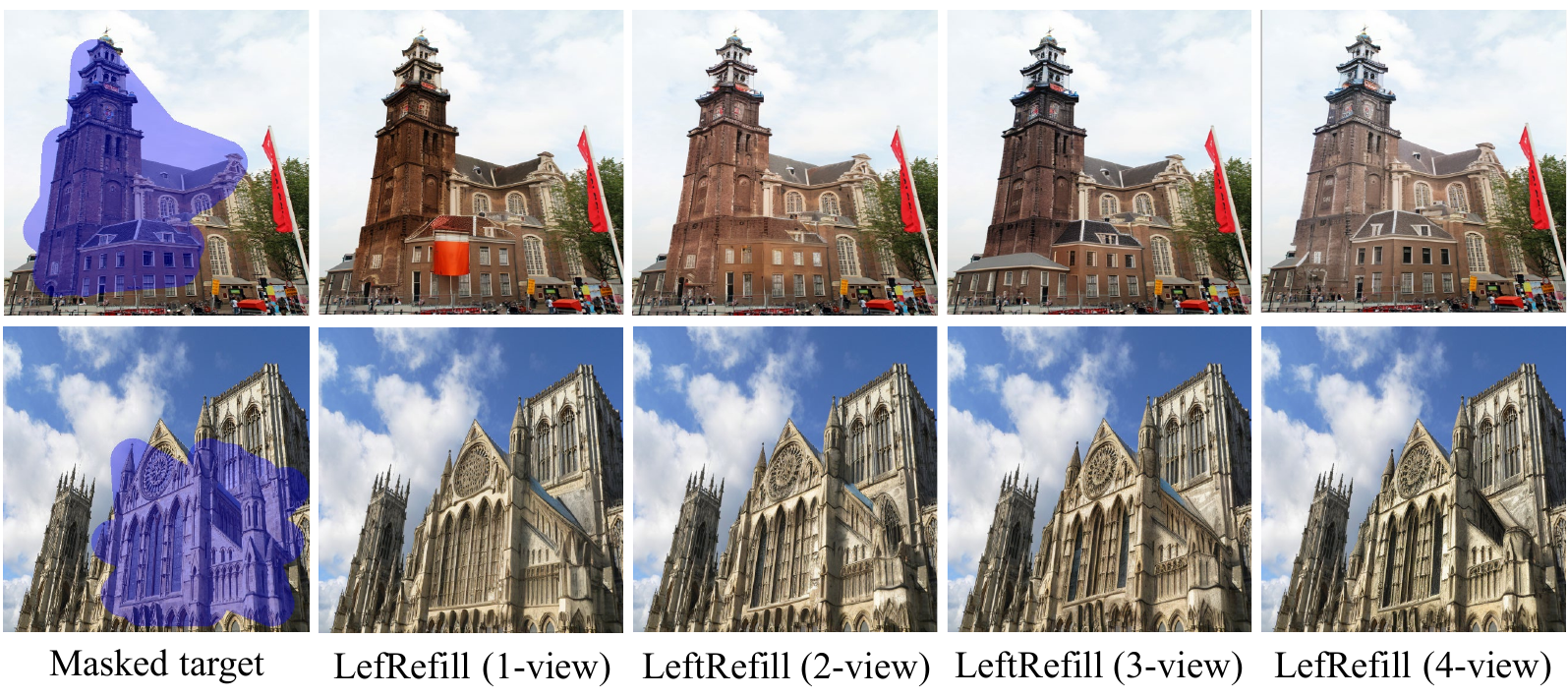}
\vspace{-0.3in}
\end{center}
   \caption{ Multi-view Ref-inpainting qualitative results. Increasing the reference view number improves the quality of repaired targets. 
   \label{fig:multview_ref_inpainting}}
\vspace{-0.3in}
\end{figure}

\noindent\textbf{Implementation Details.}
Our LeftRefill is based on the inpainting fine-tuned SD~\citep{rombach2022high} with 0.8 billion parameters.
For the task and view prompt tuning, there are 50 trainable prompt tokens at all.
We use 90\% tokens to represent the task embeddings, while 10\% tokens indicate each view respectively. 
We use the AdamW optimizer with a weight decay of 0.01 to optimize LeftRefill. 
For the Ref-inpainting, the prompt tuning's learning rate is 3e-5. Moreover, 75\% masks are randomly generated, and 25\% of them are matching-based masks.
For the NVS, LeftRefill could be tested with the adaptive masking strengthened by the foreground segmentation model.
Concretely, we first enlarge the reference segmentation mask and generate a coarse target with fewer DDIM steps.
We extract the new mask from the coarse target and then further randomly enlarge it as the final target mask.
The NVS LeftRefill is trained with 512 batch size and learning rate 3e-5.
We show that a simplified LeftRefill with just 48 batch size can also be converged for NVS, which breaks through the limitation of Zero123~\cite{liu2023zero}.
More details about matching-based masks, adaptive masking, and training schedules are discussed in the supplementary.

\begin{figure}[htb]
\begin{center}
\includegraphics[width=0.9\linewidth]{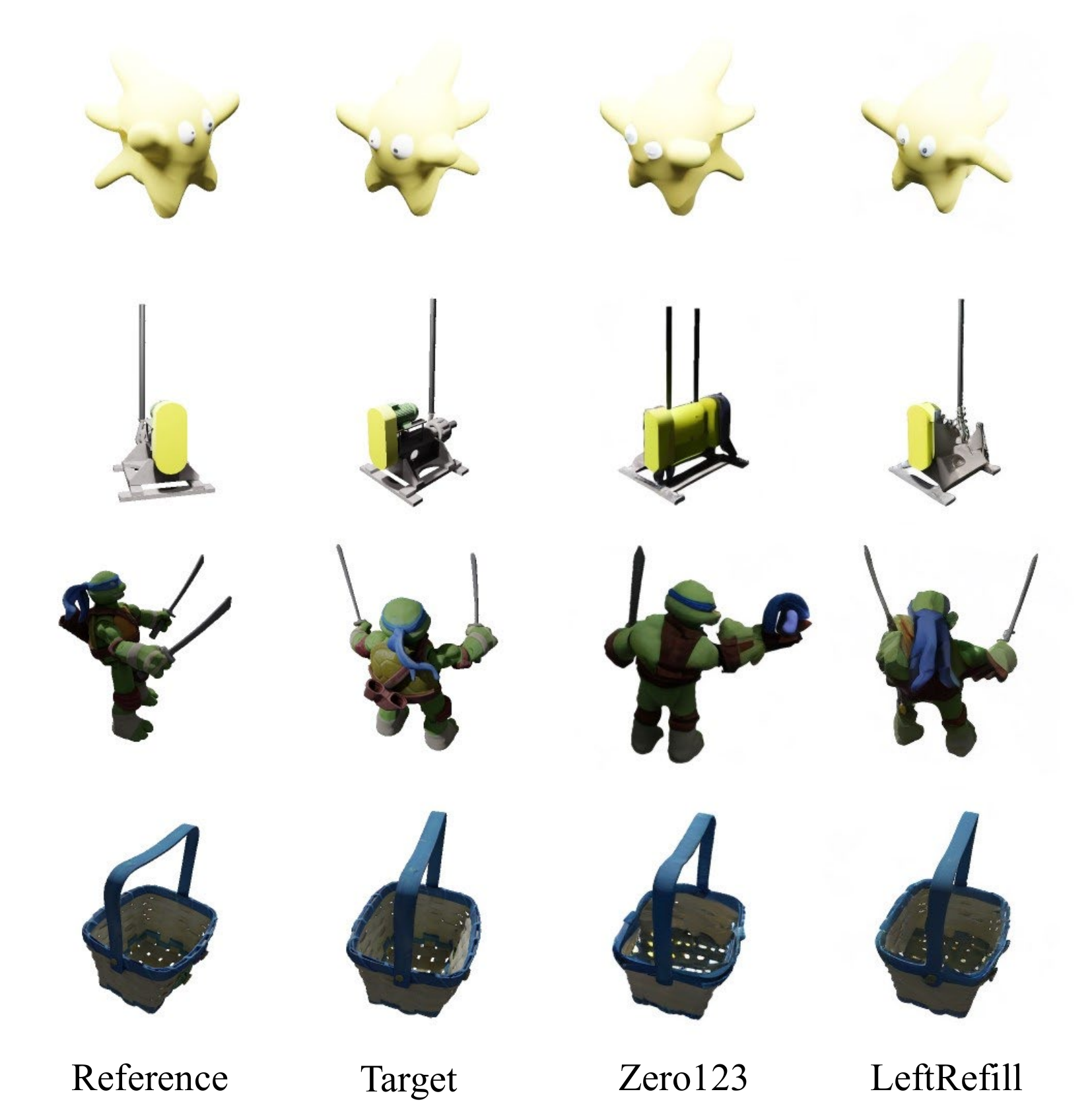}
\vspace{-0.15in}
\end{center}
   \caption{  
   NVS results on Objaverse~\citep{deitke2022objaverse} (row1, 2) and Google Scanned Objects~\citep{downs2022google} (row3, 4) from a single reference image. 
   \label{fig:nvs_obj}}
\vspace{-0.15in}
\end{figure}

\subsection{Results of Reference-guided Inpainting}
\label{sec:exp_ref_inpainting}

\noindent\textbf{Results of One-view Reference.}
We thoroughly compared the specific Ref-inpainting method~\citep{zhou2021transfill} and existing image reference-based variants of SD with one-view reference in \Cref{tab:reference_inpainting_quantitative} and \Cref{fig:reference_guided_qualitative}. 
Note that ControlNet~\citep{zhang2023adding} fails to learn the correct spatial correlation between reference images and masked targets, even enhanced with trainable cross-attention learned between reference and target features. 
Furthermore, we try to warp ground-truth latent features with image matching~\citep{tang2022quadtree} as the reference guidance for ControlNet, but the improvement is not prominent.
Perceiver~\citep{jaegle2021perceiver} and Paint-by-Example~\citep{yang2023paint} align and learn image features from Image CLIP. Since image features from CLIP contain high-level semantics, they fail to deal with the fine-grained Ref-inpainting as shown in \Cref{fig:reference_guided_qualitative}(e)(f).
Though TransFill~\citep{zhou2021transfill} achieves proper results in PSNR and SSIM, it suffers from blur and color difference as in \Cref{fig:reference_guided_qualitative}(g) with challenging viewpoints.
LeftRefill enjoys substantial advantages in both qualitative and quantitative comparisons with negligible trainable weights. 
Particularly, spatially stitching reference and target views together achieves consistent improvements.
Thus it is intuitive and convincing that all U-net modules contribute to improved inpainting results with a whole stitched image, as opposed to using only attention modules.
We further compare LeftRefill with TransFill on their officially provided real-world dataset in the supplementary.

\noindent\textbf{Results of Multi-view Ref-inpainting.} 
We verified models trained with different numbers of reference views in \Cref{tab:reference_inpainting_quantitative} (lower).
As the number of reference views increases, there is a notable enhancement in inpainting capability. As qualitatively compared in \Cref{fig:multview_ref_inpainting}, more consistent references lead to robust inpainting results with sensible structures.

\subsection{Results of Novel View Synthesis}
\label{sec:exp_nvs}

\begin{figure}
\begin{center}
\includegraphics[width=0.9\linewidth]{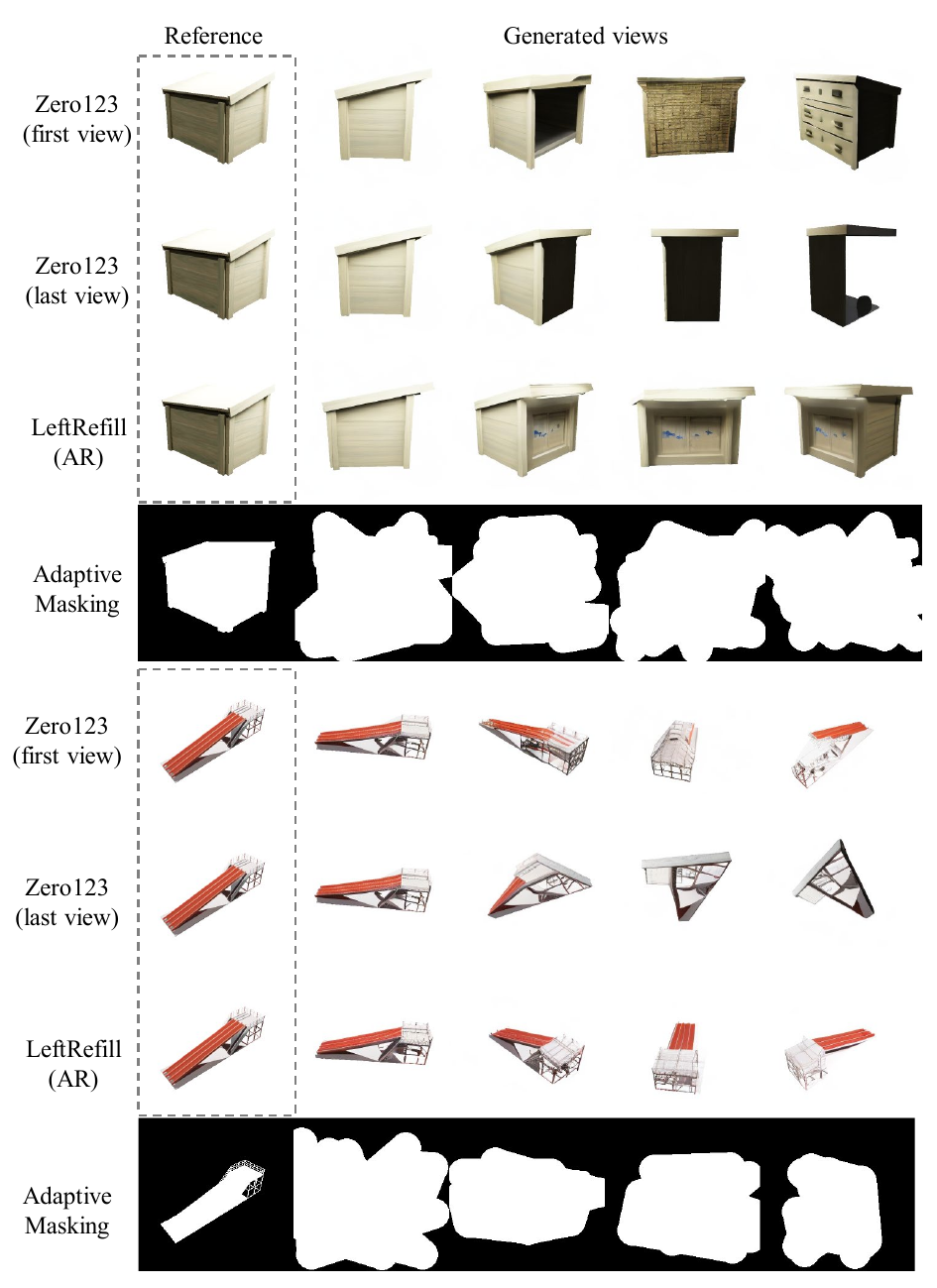}
\vspace{-0.2in}
\end{center}
   \caption{ Sequential generative results from a single view. Zero123~\citep{liu2023zero} are conditioned on real reference (first view) and last generated view (last view), while LeftRefill is based on AR.
   \label{fig:ar_nvs}}
\vspace{-0.2in}
\end{figure}

\noindent\textbf{Results of Single-view NVS.}
We compare the quantitative NVS results on Objaverse in \Cref{tab:novel_view_quantitative}.
The qualitative results for both Objaverse and Google Scanned Objects are compared in \Cref{fig:nvs_obj}.
Without specific annotations, all NVS results are based on adaptive masking.
The CLIP score~\citep{radford2021learning} is compared to evaluate the similarity between the generation and the target.
Specifically, LeftRefill fine-tuned with the whole LDM enjoys substantial achievements compared to the state-of-the-art competitor Zero123~\cite{liu2023zero}, even though Zero123 might have seen our validation in Objaverse. 
Moreover, the LoRA-based LeftRefill~\cite{hu2021lora} is still competitive with a very moderate training setting (batch size 48 with 2 A6000 GPUs) in \Cref{tab:novel_view_quantitative}.
We further compare the training log of LeftRefill and Zero123 in the supplementary to investigate the training convergence for NVS.
The contextual inpainting-based LeftRefill enjoys a substantially faster convergence and superior image quality.
So LeftRefill enjoys a good balance between training efficiency and performance.



\begin{table}
\small 
\vspace{-0.05in}
\caption{
Results of 1-view NVS conditioned on different numbers of reference views on Objaverse~\citep{deitke2022objaverse}. `w.o. AM' indicates NVS results without Adaptive Masking (AM).
\label{tab:novel_view_quantitative}}
\vspace{-0.1in}
\centering
\renewcommand\tabcolsep{2pt}
\begin{tabular}{lccccc}
\toprule 
Methods & Ref-View & PSNR$\uparrow$ & SSIM$\uparrow$ & LPIPS$\downarrow$ & CLIP$\uparrow$\tabularnewline
\midrule 
Zero123~\citep{liu2023zero} & 1 & 19.402 & 0.858 & 0.1309 & 0.7816 \tabularnewline
LeftRefill (LoRA) & 1 & 19.514 & 0.869 & 0.1534 & 0.7589\tabularnewline
\textcolor{gray}{LeftRefill (w.o. AM)} & \textcolor{gray}{1} & \textcolor{gray}{21.675} & \textcolor{gray}{0.887} & \textcolor{gray}{0.1089} & \textcolor{gray}{{0.7959}}\tabularnewline
LeftRefill & 1 & \textbf{21.404} & \textbf{0.882} & \textbf{0.1151} & \textbf{0.7972}\tabularnewline
\midrule 
LeftRefill & 2 & 22.935 & 0.895 & 0.0871 & 0.8280\tabularnewline
LeftRefill & 3 & 24.107 & 0.908 & 0.0722 & 0.8432\tabularnewline
LeftRefill & 4 & \textbf{24.685} & \textbf{0.911} & \textbf{0.0634} & \textbf{0.8495}\tabularnewline
\bottomrule
\end{tabular}
\vspace{-0.15in}
\end{table}

\begin{table}
\small 
\caption{
Results of 4-view NVS generations based on 1 reference view on Objaverse~\citep{deitke2022objaverse}.
P-CLIP means pairwise CLIP score showing consistency of generated views.
The reference (Ref) can be categorized into the first ground-truth view and the last generated view, while we also provide the AR results of LeftRefill.
\label{tab:one_to_more_nvs}}
\vspace{-0.1in}
\centering
\renewcommand\tabcolsep{3pt}
\begin{tabular}{lcccccc}
\toprule 
Methods & Ref & PSNR$\uparrow$ & SSIM$\uparrow$ & LPIPS$\downarrow$ & CLIP$\uparrow$ & P-CLIP$\uparrow$\tabularnewline
\midrule
Zero123 & First & 19.265 & 0.855 & 0.1366 & 0.7723 & 0.7756\tabularnewline
Zero123 & Last & 14.621 & 0.767 & 0.2569 & 0.6921& 0.7667\tabularnewline
LeftRefill & First & \textbf{21.573} & \textbf{0.883} & \textbf{0.1143} & \textbf{0.7964} & 0.7709 \tabularnewline
LeftRefill & AR & 21.271 & 0.882 & 0.1195 & 0.7882 & \textbf{0.7958}\tabularnewline
\bottomrule
\end{tabular}
\vspace{-0.15in}
\end{table}

\begin{figure}
\begin{center}
\includegraphics[width=0.95\linewidth]{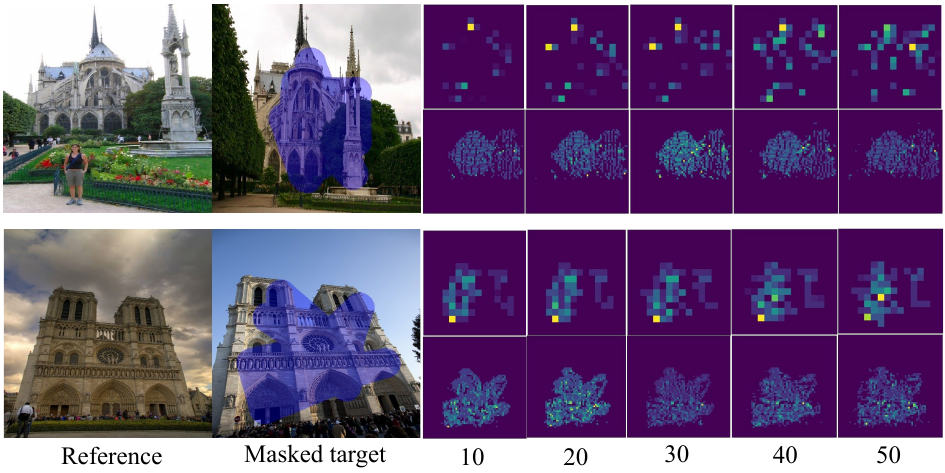}
\vspace{-0.2in}
\end{center}
   \caption{ Visualization of attention scores in LeftRefill for Ref-inpainting across different DDIM steps. We show scores from reference views attended by masked regions. The upper row shows attention scores from the 8th self-attention (1/32 scale), while the bottom row shows ones from the 14th self-attention (1/8 scale). 
   \label{fig:attn_analysis}}
\vspace{-0.1in}
\end{figure}

\begin{table*}[h]
\small 
\caption{ Ablation studies for the setting of prompt tuning in Ref-inpainting. 
Left: `Shallow' means only prompt tuning to text embedding, while `Deep' indicates tuning additional embedding features to different cross-attention layers (16 layers) in SD. 
Right: validating the influence of the length of shared (Task) and unshared (View) prompts with 3-view Ref-inpainting.
\label{tab:referece_inpainting_pt_len_ablation}}
\vspace{-0.1in}
\centering
\begin{tabular}{cc}
\renewcommand\tabcolsep{2pt}
\begin{tabular}{cccccc}
\toprule 
Prompt Type & Length & PSNR$\uparrow$ & SSIM$\uparrow$ & LPIPS$\downarrow$ & Params\tabularnewline
\midrule 
Shallow & 25 & 20.35 & 0.827 & 0.104 & +0.025M\tabularnewline
Shallow & 50 & \textbf{20.49} & 0.829 & \textbf{0.103} & +0.05M\tabularnewline
Shallow & 75 & 20.38 & \textbf{0.830} & 0.104 & +0.075M\tabularnewline
Deep ($\times$16) & 25(400) & 20.15 & 0.825 & 0.106 & +0.4M\tabularnewline
\bottomrule 
\end{tabular} & %
\renewcommand\tabcolsep{1.75pt}
\begin{tabular}{cccccc}
\toprule 
Task & View & PSNR$\uparrow$ & SSIM$\uparrow$ & LPIPS$\downarrow$ & Params\tabularnewline
\midrule 
50 & 0 & 21.224 & 0.838 & 0.0941 & +0.05M\tabularnewline
45 & 5 & \textbf{21.356} & \textbf{0.840} & \textbf{0.0901} & +0.06M\tabularnewline
25 & 25 & 21.127 & 0.836 & 0.0950 & +0.11M\tabularnewline
5 & 45 & 20.744 & 0.832 & 0.1040 & +0.14M\tabularnewline
0 & 50 & 20.563 & 0.831 & 0.1110 & +0.15M\tabularnewline
\bottomrule 
\end{tabular}\tabularnewline
\end{tabular}
\vspace{-0.15in}
\end{table*}

\noindent\textbf{Results of Multi-view NVS.}
Quantitative results of multi-view NVS are shown in the lower of \Cref{tab:novel_view_quantitative}.
Obviously, more reference views lead to better reconstruction quality of LeftRefill.
Moreover, additional reference images could substantially alleviate the ambiguity, improving the final results with consistent geometry.
Benefited by AR, LeftRefill can be also generalized to synthesize a group of consistent images with different viewpoints from a single view as shown in \Cref{fig:ar_nvs} and \Cref{tab:one_to_more_nvs}. 
We introduce the pairwise CLIP score (P-CLIP) to verify the consistency of all generated samples. 
LeftRefill outperforms Zero123 in most metrics, while AR could prominently improve the consistency with just a little quality degradation.
Our method can also be generalized to real-world data as shown in the supplementary.

\subsection{Analysis and Ablation Studies}
\label{sec:ablation}

\noindent\textbf{Self-Attention Analysis.}
We show the visualization of self-attention scores attended by masked regions for Ref-inpainting across different DDIM steps in \Cref{fig:attn_analysis}. 
Self-attention can capture correct feature correlations without any backbone fine-tuning. 
As diffusion sampling progresses, self-attention modules gradually shift their focus from specific key points to broader related regions, which is convincing and intuitive.
Because the key landmarks help to swiftly locate the spatial correlation between the reference and target, while the extended receptive fields further refine the generation for the following sampling steps. 
More analysis about the attention visualization with increased reference views is shown in the supplementary.

\begin{table}
\vspace{-0.1in}
\small 
\caption{ NVS results of LeftRefill-simple with different reference views with/without incremental Positional Encoding (PE).
\label{tab:PE_ablation}}
\vspace{-0.1in}
\centering
\begin{tabular}{ccccc}
\toprule 
Ref-View & PE & PSNR$\uparrow$ & SSIM$\uparrow$ & LPIPS$\downarrow$\tabularnewline
\midrule
1 & $\times$ & 20.352 & 0.873 & 0.132\tabularnewline
1 & $\checkmark$ & \textbf{20.508} & \textbf{0.875} & \textbf{0.128}\tabularnewline
\midrule
4 & $\times$ & 22.097 & 0.888 & 0.099\tabularnewline
4 & $\checkmark$ & \textbf{22.324} & \textbf{0.890} & \textbf{0.095}\tabularnewline
\bottomrule 
\end{tabular}
 \end{table}

\noindent\textbf{Prompt Settings.} 
The length and depth used in the task and view prompt tuning are explored in \Cref{tab:referece_inpainting_pt_len_ablation}. Different from~\cite{jia2022visual}, we find that LeftRefill is relatively robust in the length selection. Thus we select 50 for both Ref-inpainting and NVS.
Moreover, the deep prompt with much more trainable prompts for different cross-attention layers does not perform well, which may suffer from a little overfitting. 
For the multi-view scene, we empirically evaluate the 3-view-based Ref-inpainting performance with various proportions of task\&view prompt lengths in the right of \Cref{tab:referece_inpainting_pt_len_ablation}. 
Increasing the proportion of view tokens initially improves the results, followed by a subsequent decline. 
We think that a few unshared view tokens contribute valuable view orders, while too many unshared tokens would increase the learning difficulty, leading to an inferior prompt tuning performance.

\noindent\textbf{Incremental Positional Encoding.} 
We incrementally add the concatenation of learnable view embedding and absolute positional encoding to each attention block for NVS (\Cref{eq:PE}), improving the performance of both single-view and multi-view-based NVS as verified in \Cref{tab:PE_ablation}.

\begin{figure}
\begin{center}
\includegraphics[width=0.95\linewidth]{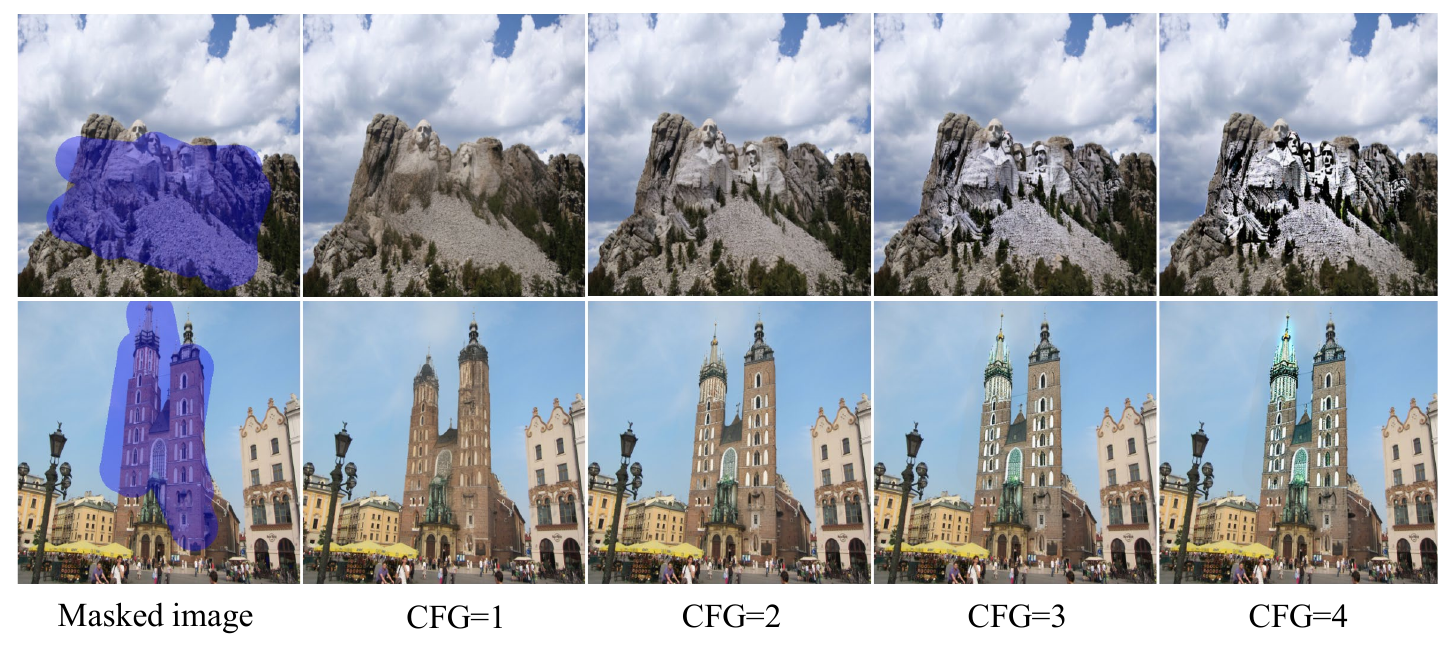}
\vspace{-0.25in}
\end{center}
   \caption{Ref-inpainting results with different CFG weights, which take a trade-off between structural and textural recoveries.
   \label{fig:ref_inpainting_cfg}}
\vspace{-0.15in}
\end{figure}

\noindent\textbf{Classifier-Free Guidance (CFG)~\citep{ho2022classifier}.}
We find that CFG can enhance the performance of Ref-inpainting even without training with prompt dropout as in \Cref{fig:ref_inpainting_cfg}. 
The adjusting of CFG could be seen as the trade-off between structural and textural recoveries. High CFG scales lead to over-saturated results with superior structure. We empirically set CFG to 2.0 and 2.5 for Ref-inpainting and NVS respectively. More quantitative results and details about the CFG setting of NVS are discussed in the supplementary.


\section{Conclusion}

In this paper, we propose LeftRefill, formulating reference-based synthesis as inpainting tasks and addressing them end-to-end as a human painter. Benefiting from the prompt tuning and the well-learned attention modules in large T2I models, LeftRefill can address the spatially sophisticated Ref-inpainting and NVS efficiently. Moreover, LeftRefill could be easily extended to tackle multi-view generation tasks. We also propose block casual masking to accomplish NVS with consistent results autoregressively.
Comprehensive experiments on Ref-inpainting and NVS show the effectiveness and efficiency of LeftRefill.

{
    \small
    \bibliographystyle{ieeenat_fullname}
    \bibliography{main}
}

\appendix

\section{Broader Impacts}

This paper exploited image synthesis with text-to-image models. Because of their impressive generative abilities, these models may produce misinformation or fake images. 
So we sincerely remind users to pay attention to it.
Besides, privacy and consent also become important considerations, as generative models are often trained on large-scale data.
Furthermore, generative models may perpetuate biases present in the training data, leading to unfair outcomes. 
Therefore, we recommend users be responsible and inclusive while using these text-to-image generative models.
Note that our method only focuses on technical aspects. Both images and pre-trained models used in this paper are all open-released.

\begin{figure*}[htb!]
\begin{center}
\includegraphics[width=1.0\linewidth]{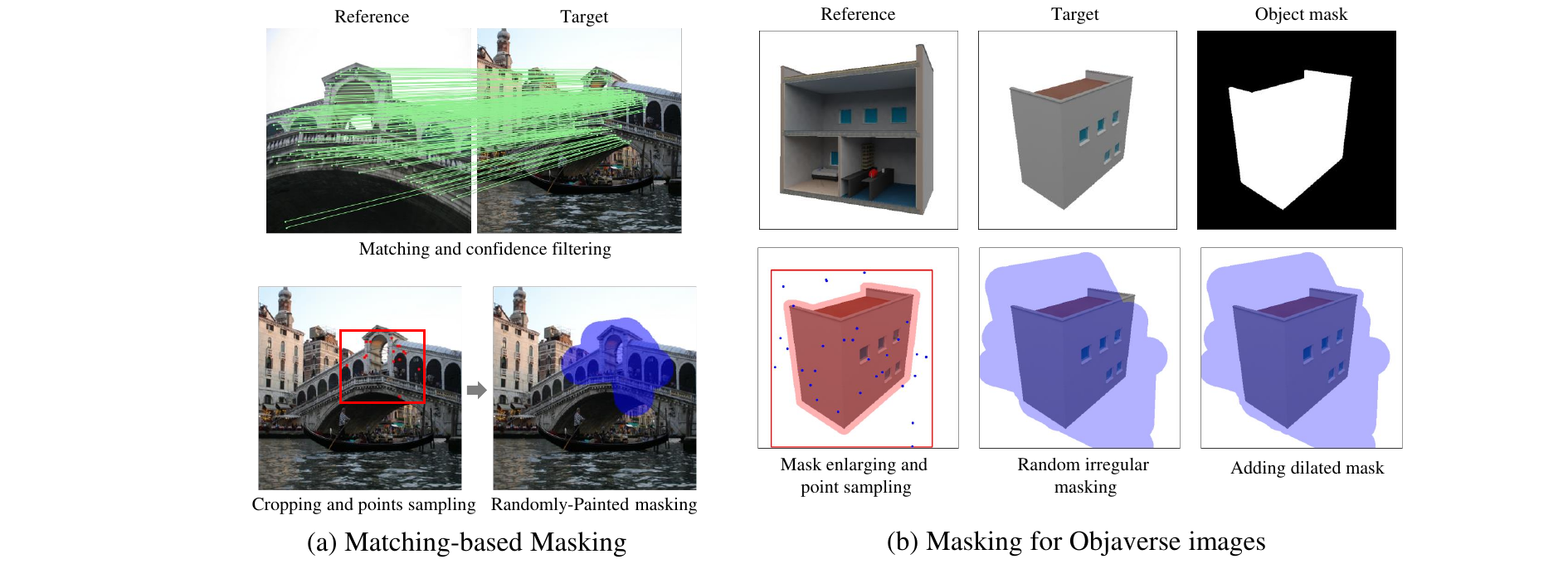}
\vspace{-0.15in}
\end{center}
   \caption{ The illustration of (a) matching-based masking for Ref-inpainting, and (b) masking strategy used for NVS on Objaverse~\citep{deitke2022objaverse}.
   \label{fig:data_process}}
\vspace{-0.15in}
\end{figure*}

\section{Data Processing and Implementation Details}
\label{sec:data_processing}

\subsection{Data Processing for Ref-inpainting}
\label{sec:matching_mask}

\noindent\textbf{Matching-based Masking.}
For the Ref-inpainting, we find that the widely used irregular mask~\citep{dong2022incremental,zhou2021transfill,zhao2022geofill} fails to reliably evaluate the capability of spatial transformation and structural preserving. Therefore, as shown in \Cref{fig:data_process}(a), we propose the matching-based masking method.
Specifically, we first utilize the scene info provided by MegaDepth~\citep{li2018megadepth} to select out the image pairs which have an overlap rate between 40\% and 70\% Second, for each image pair, we use a feature matching model~\citep{tang2022quadtree} to detect matching key-points between the images and assign each key-points pair a confidence score. 
Next, we filter out those pairs with low confidence scores with the threshold of 0.8. Then we randomly crop a 20\% to 50\% sub-space in the matched region and sample 15 to 30 key points as vertices to be painted across for the final masks. 
The matching-based mask not only improves the reliability during the evaluation but also facilitates the performance in the training phase as in \Cref{tab:reference_inpainting_config_ablation}.

We split 505 pairs from MegaDepth~\citep{li2018megadepth} as the validation, including some manual masks from ETH3D scenes~\citep{schops2017multi}.
For the multi-view testing set, we further filter all scenes and retain the ones with at least 4 reference views. Thus there are 482 images in the final multi-view testing set.

\subsection{Data Processing for NVS}
For the NVS, we first dilate the object mask and randomly sample points in the enlarged mask bounding box to paint the irregular mask. Then, we unite the dilated object mask to completely cover target images as in \Cref{fig:data_process}(b).
We find that local masking is still very important for fast convergence and stable fine-tuning as empirically verified in experiments.
For the data processing on Objaverse~\citep{deitke2022objaverse}, Zero123~\citep{liu2023zero} provided images including 800k various scenes with object masks. 
For each scene, 12 images are rendered in 256$\times$256 with different viewpoints.
Following~\cite{liu2023zero}, the spherical coordinate system is used to convert the relative pose $\Delta p$ into the polar angle $\theta$, azimuth angle $\phi$, and radius $r$ distanced from the canonical center as $\Delta p=(\Delta\theta,\mathrm{sin}\Delta\phi,\mathrm{cos}\Delta\phi,\Delta r)$, where the azimuth angle is sinusoidally encoded to address the non-continuity.
In practice, we calculate the relative pose between the \emph{first} view and the target view for the pose input to LeftRefill.
For example, given a group of 4-view stitched input images, we provide relative poses of view 0-to-1, 0-to-2, 0-to-3, and 0-to-4, respectively.
For the masking of Objaverse images, we dilate the object mask and related bounding box with 10 to 25 kernel size and 5\% to 20\% respectively. Then we randomly sample 20 to 45 points to paint the irregular masks.

We select 500 scenes from Objaverse as the validation, while others are used as the training set. Note that there exists an overlap between our validation and Zero123's training set~\citep{liu2023zero}, but our method still outperforms the official Zero123 as in the main paper.

\subsection{Training Details}
\label{sec:training_details_supp}

\begin{table*}
\small 
\caption{ Training details of LeftRefill. NVS (4-view) and Ref-inpainting (4-view) are trained on $\times$8 and $\times$4 A800 GPUs respectively, while others are trained on $\times$2 A6000 GPUs. NVS (4-view) is fine-tuned based on NVS (1-view).
\label{tab:training_details}}
\centering
\begin{tabular}{lcccc}
\toprule
\multicolumn{1}{c}{\multirow{2}{*}{Task}} & \multicolumn{1}{c}{\multirow{2}{*}{Batch size}} & \multicolumn{2}{c}{Learning rate} & \multicolumn{1}{c}{\multirow{2}{*}{Steps}}\tabularnewline
 &  & \multicolumn{1}{c}{Prompt\&LoRA} & \multicolumn{1}{c}{Backbone} & \tabularnewline 
\midrule
Ref-inpainting (1-view) & 16 & 3e-5 & / & 6k \tabularnewline
Ref-inpainting (2-view) & 16 & 3e-5 & / & 6k \tabularnewline
Ref-inpainting (3-view) & 24 & 3e-5 & / & 16k \tabularnewline
Ref-inpainting (4-view) & 64 & 5e-5 & / & 16k \tabularnewline
NVS-simple (1-view) & 48 & 1e-4 & 1e-5 & 80k \tabularnewline
NVS (4-view) & 512 & 1e-4 & 3e-5 & 110k \tabularnewline
\bottomrule
\end{tabular}
\end{table*}

\begin{figure}
\begin{center}
\includegraphics[width=1.0\linewidth]{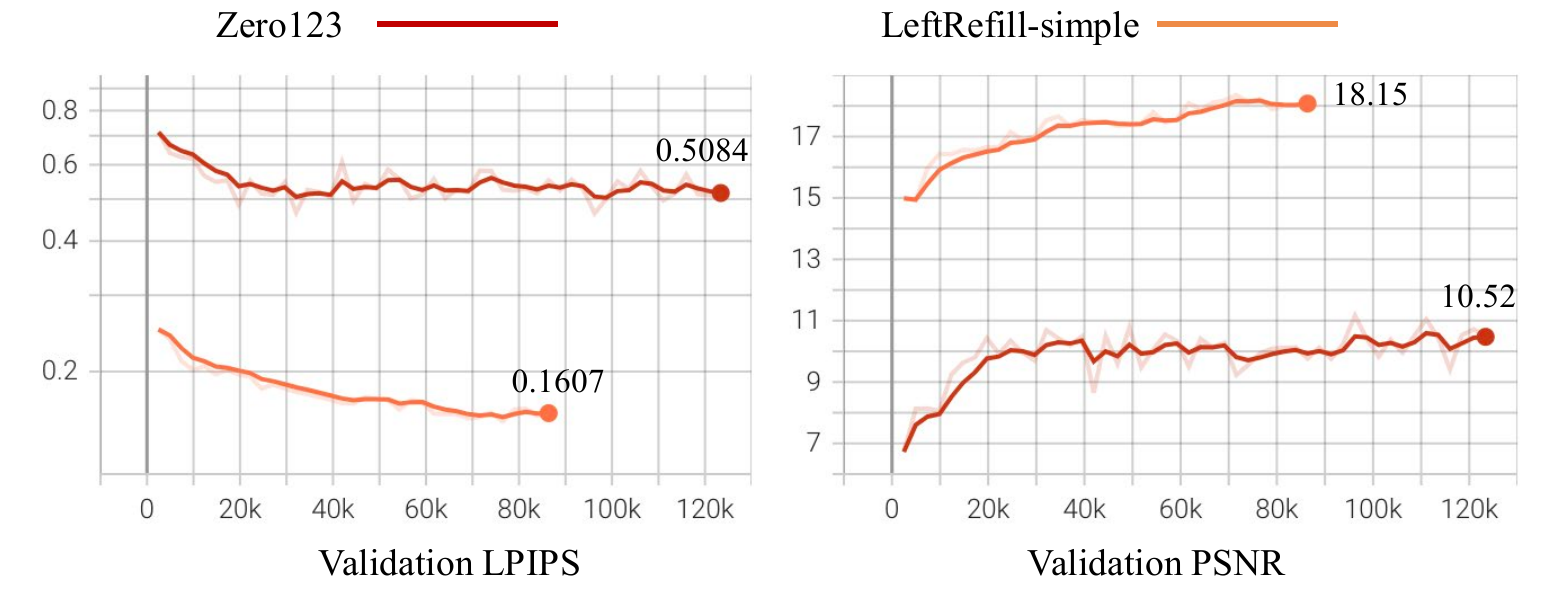}
\vspace{-0.2in}
\end{center}
   \caption{ NVS training logs of LeftRefill and Zero123~\citep{liu2023zero} on Objaverse~\citep{deitke2022objaverse} (batch size 48, learning rate 1e-5).
   \label{fig:train_lpips_psnr}}
\vspace{-0.15in}
\end{figure}

We show the training details in \Cref{tab:training_details}. 
LeftRefill is efficient in being trained for various tasks. 
To further demonstrate the effectiveness of LeftRefill, we provide the training log of LeftRefill and Zero123 in \Cref{fig:train_lpips_psnr}. 
Obviously, the contextual inpainting-based LeftRefill enjoys a substantially faster convergence and superior performance.

\begin{table}
\small 
\vspace{-0.05in}
\caption{
Results of 1-view NVS on Objaverse. 
Zero123* was re-trained with the same setting as LeftRefill-simple (batch 48).
\label{tab:NVS_limited_set}}
\centering
\renewcommand\tabcolsep{1.8pt}
\begin{tabular}{lccccc}
\toprule 
Methods & PSNR$\uparrow$ & SSIM$\uparrow$ & LPIPS$\downarrow$ & CLIP$\uparrow$\tabularnewline
\midrule  
Zero123{*} (re-trained) & 14.316 & 0.802 & 0.3455 & 0.6549\tabularnewline
LeftRefill-simple (prompt tuning) & 16.385 & 0.855 & 0.2468 & 0.7107\tabularnewline
LeftRefill-simple (LoRA)~\citep{hu2021lora} & 19.514 & 0.869 & 0.1534 & 0.7589\tabularnewline
LeftRefill-simple (fine-tune) & \textbf{20.508} & \textbf{0.875} & \textbf{0.1288} & \textbf{0.7763}\tabularnewline
\bottomrule
\end{tabular}
\vspace{-0.15in}
\end{table}

\begin{algorithm*}[htb]
  \caption{ Pseudo codes for block casual masking.}
  \label{alg:block_casual_masking}
    \definecolor{codeblue}{rgb}{0.25,0.5,0.5}
    \lstset{
      basicstyle=\fontsize{7.5pt}{7.5pt}\ttfamily\bfseries,
      commentstyle=\fontsize{7.5pt}{7.5pt}\color{codeblue},
      keywordstyle=\fontsize{7.5pt}{7.5pt},
    }
\vspace{-0.1in}
\begin{lstlisting}[language=python]
# view: the view number
# length: length of the sequence, usually be h*w

mask = zeros((view, length)) # [view,length]
mask[:, 0] = 1
mask = cumsum(mask.reshape(1, view * length), dim=1) # [1,view*length]
mask = (mask.T >= mask).float()  # [view*length,view*length]
mask = 1 - mask # masked regions are 1, unmasked regions are 0
mask = mask.masked_fill(mask == 1, -inf) # let all masked regions to -inf
\end{lstlisting}
\vspace{-0.1in}
\end{algorithm*}

\begin{algorithm*}[htb]
  \caption{ Pseudo codes for the attention visualization.}
  \label{alg:attention_score}
    \definecolor{codeblue}{rgb}{0.25,0.5,0.5}
    \lstset{
      basicstyle=\fontsize{7.5pt}{7.5pt}\ttfamily\bfseries,
      commentstyle=\fontsize{7.5pt}{7.5pt}\color{codeblue},
      keywordstyle=\fontsize{7.5pt}{7.5pt},
    }
\vspace{-0.1in}
\begin{lstlisting}[language=python]
# x: [b,2hw,c], input feature for attention module (left:reference, right:target)
# mask: [b,2hw,1], input 0-1 mask; 1 means masked regions

q, k = matmul(x, Wq), matmul(x, Wk) # [b,2hw,c], project x to query (q) and key (k)
A = matmul(q, k.T) # [b,2hw,2hw], get attenion map
A = mean(A * mask , dim=1) # [b,2hw] get mean scores attended by masked regions
A = A.reshape(b,h,w)[:, :, :w//2] # [b,h,w], show reference attention score only
\end{lstlisting}
\vspace{-0.1in}
\end{algorithm*}

\subsection{Differences between Ref-inpainting and NVS}
\label{sec:diff_inpainting_nvs}

The proposed framework, LeftRefill, serves as a generalized solution catering to both Ref-inpainting and NVS, as detailed in our main paper. 
However, given the substantial disparities between NVS and Ref-inpainting tasks, we present a comprehensive overview of the minor distinct implementations of LeftRefill tailored for each task.
Notably, as the inpainting fine-tuned SD suffers from a large gap in tackling NVS directly, LeftRefill requires slightly more modifications to optimize its performance for NVS. The following key adjustments were identified:

\begin{itemize}
\item[1)] NVS needs to be fine-tuned for the whole LDM, while Ref-inpainting only requires prompt tuning. Note that both tasks could be addressed without test-time fine-tuning through LeftRefill.

\item[2)] NVS needs another pose FC to encode relative pose information to CLIP-H.

\item[3)] To enhance the performance of NVS, positional encoding is added before each self-attention module of LeftRefill. However, our experiments did not reveal significant improvements when positional encoding was applied to Ref-inpainting.

\item[4)] The self-attention module of multi-view NVS should be processed with the block casual masking strategy for autoregressive generation. 
In contrast, multi-view Ref-inpainting does not require autoregressive generation since only one view needs to be generated.
\end{itemize}

Despite these nuanced differences between Ref-inpainting and NVS within the LeftRefill framework, we clarify that it remains a sufficiently generalized model capable of effectively handling reference-based synthesis.

\section{Autoregressively Sequential Generation}
\label{sec:supp_sequential_gen}

\begin{figure}
\begin{center}
\includegraphics[width=1.0\linewidth]{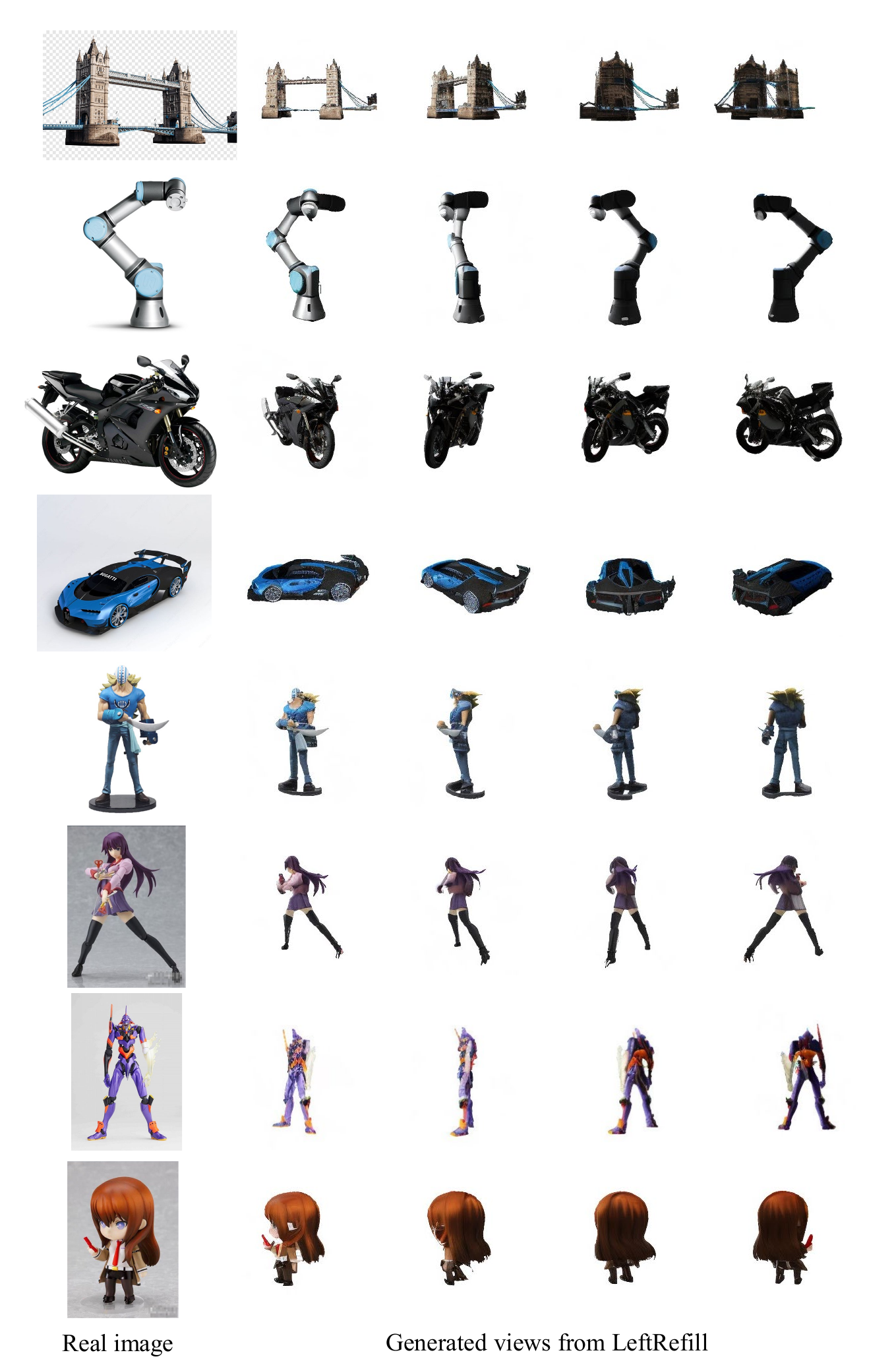}
\vspace{-0.1in}
\end{center}
   \caption{ 
   Consistent real-world NVS results generated by LeftRefill. 
   \label{fig:real_world_cases}}
\vspace{-0.1in}
\end{figure}

\begin{figure*}
\begin{center}
\includegraphics[width=1.0\linewidth]{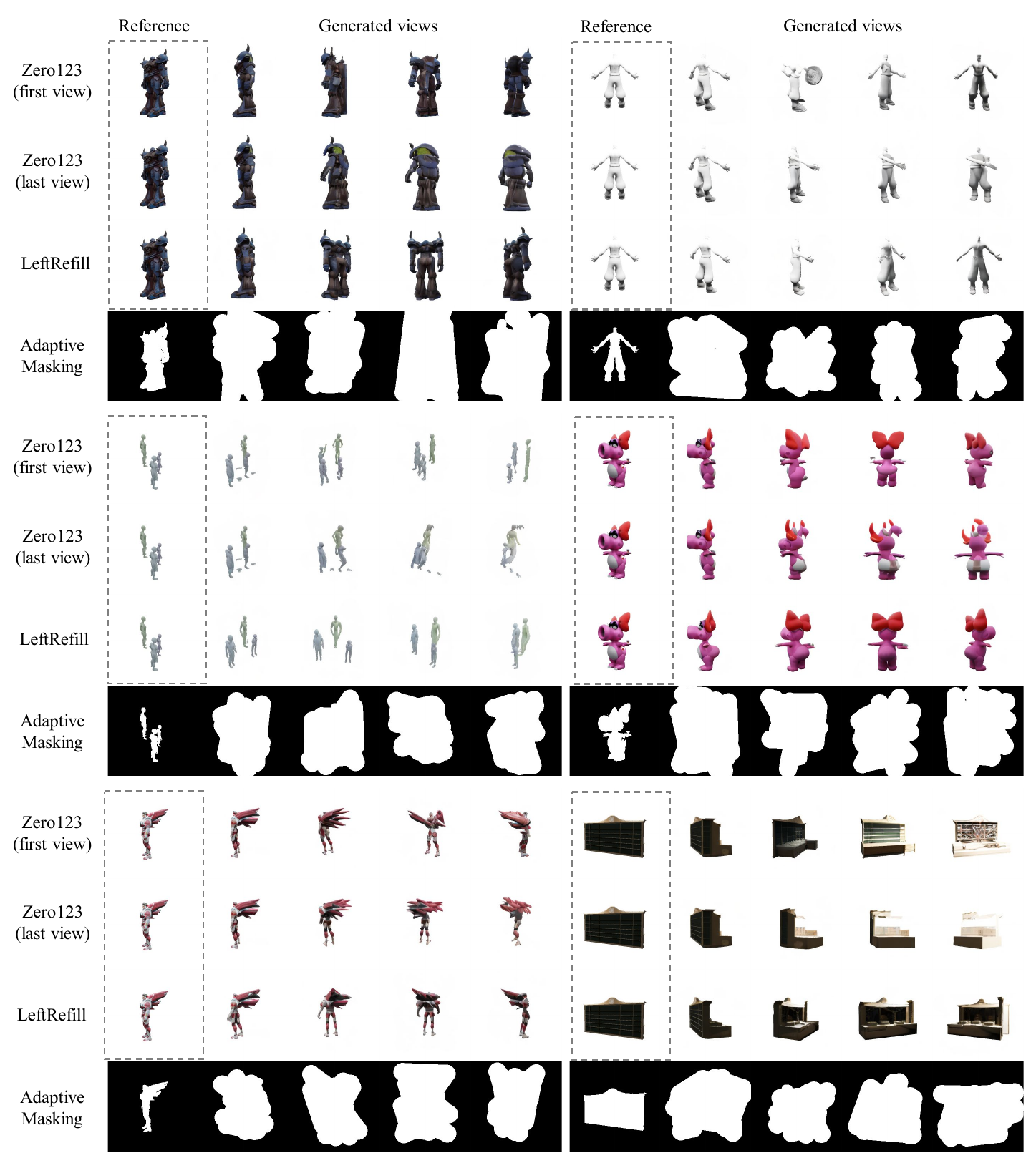}
\vspace{-0.1in}
\end{center}
   \caption{ The sequential generative results from a single view. Zero123's~\citep{liu2023zero} results are conditioned on the real reference (first view) and the last generated view (last view) respectively.
   \label{fig:ar_nvs_supp}}
\vspace{-0.1in}
\end{figure*}

To verify the generalization of our method, we generate more groups of multi-view images through a single input view as in \Cref{fig:ar_nvs_supp}. Moreover, we test several real-world cases with one RGB input in \Cref{fig:real_world_cases}. All poses are initialized to [$0.5\pi, 0, 1.5$] for polar angle, azimuth angle, and radius distance, respectively. The proposed LeftRefill can be well generalized to real-world cases.

\subsection{Adaptive Masking} 
\label{sec:adaptive_masking}
\begin{figure*}[htb!]
\begin{center}
\includegraphics[width=0.95\linewidth]{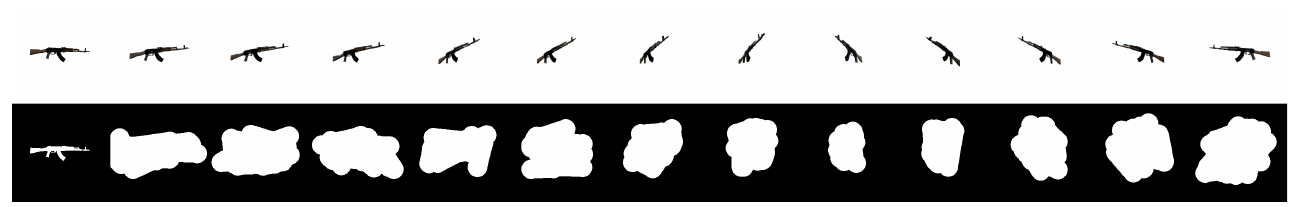}
\vspace{-0.1in}
\end{center}
   \caption{ Long sequence synthesis from a single image (upper) with adaptive masking (bottom). The leftmost image and mask are the input while others are generated.
   \label{fig:nvs_long_seq}}
\vspace{-0.1in}
\end{figure*}

One may ask that the masking strategy used in \Cref{fig:data_process}(b) suffers from shape leakages, which lead to unreliable metrics in the main paper. We should clarify that our method can perform well only with the reference mask, which is easy to get by the salient object detection~\citep{qin2020u2}.
Specifically, we dilate the reference mask as \Cref{fig:data_process}(b). Then, a few DDIM steps~\citep{song2020denoising} are used to generate a rough synthesis in the target view. After that, we detect the foreground mask based on the rough synthesis by~\cite{qin2020u2} and further dilate this mask for the second synthesis with full DDIM steps. The adaptive masking can be well generalized to the NVS as verified in \Cref{fig:nvs_long_seq}. 
All testing results in this paper without specific descriptions are already based on adaptive masking.
Besides, we think that providing target masks according to the distance and direction priors manually is also convincing to address the challenging single-view NVS.

\begin{table*}[htb]
\small 
\caption{ Ablation studies of Ref-inpainting on MegaDepth. Left: effects of matching-based masks and inference noise $\eta$. Right: effects of different prompt initialization.
\label{tab:reference_inpainting_config_ablation}
}
\centering
\begin{tabular}{cc}
\begin{tabular}{lccc}
\toprule
Configuration & PSNR$\uparrow$ & SSIM$\uparrow$ & LPIPS$\downarrow$\tabularnewline
\midrule
baseline & 20.489 & 0.829 & 0.1029\tabularnewline 
+ Match mask & 20.574 & 0.830 & 0.1010\tabularnewline
+ $\eta$=1.0 & \textbf{20.993} & \textbf{0.837} & \textbf{0.0951}\tabularnewline
\bottomrule
\end{tabular} & %
\begin{tabular}{lccc}
\toprule
Prompt init & PSNR$\uparrow$ & SSIM$\uparrow$ & LPIPS$\downarrow$\tabularnewline
\midrule
Random & 20.810 & 0.832 & 0.0998\tabularnewline
Token-wise & 20.852 & 0.833 & 0.1002\tabularnewline
Token-avgs & \textbf{20.926} & \textbf{0.836} & \textbf{0.0961}\tabularnewline
\bottomrule
\end{tabular}\tabularnewline
\end{tabular}
\end{table*}

\begin{figure}
\begin{center}
\includegraphics[width=1.0\linewidth]{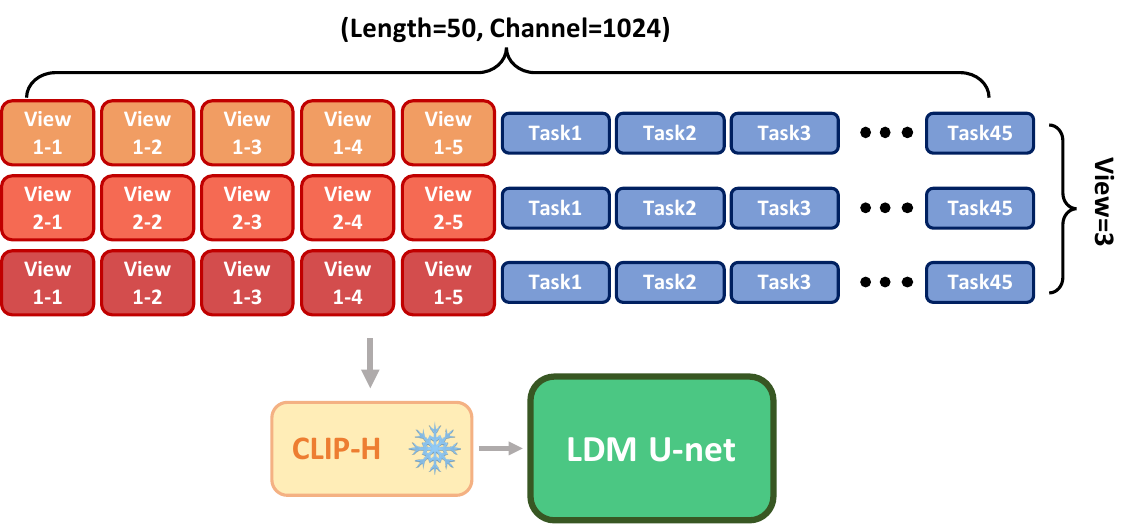}
\vspace{-0.2in}
\end{center}
   \caption{ 
   The illustration of task and view prompt tuning. 
   This case shows the situation of view number 3, the length of total prompts, unshared view prompts, and shared task prompts are 50, 5, and 45, respectively.
   \label{fig:task_prompt_tuning_visualization}}
\vspace{-0.1in}
\end{figure}

\section{Supplemental Experimental Results}

\begin{figure*}
\includegraphics[width=1.0\linewidth]{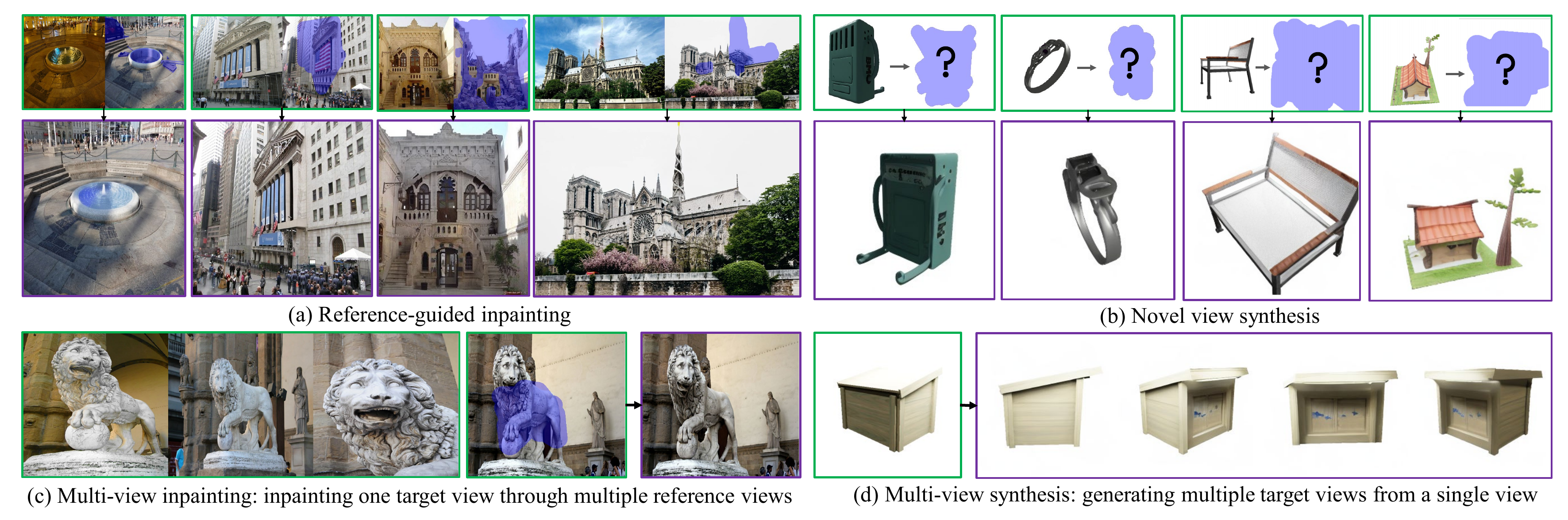}
\vspace{-0.15in}
\captionof{figure}{\small More impressive results of LeftRefill based on (a) Ref-inpainting, (b) NVS, (c) multi-view inpainting, and (d) multi-view synthesis.
   \label{fig:supple_teaser}}
\vspace{-0.15in}
\end{figure*}

We show more impressive results of LeftRefill in \Cref{fig:supple_teaser}.

\subsection{Supplemental Ablation Studies}
\label{sec:supp_ablation}

\begin{table}
\small
\centering
\caption{ Abaltions of CFG on Objaverse~\citep{deitke2022objaverse} NVS.\label{tab:cfg_nvs}}
\begin{tabular}{ccccc}
\toprule
CFG training & CFG weight & PSNR$\uparrow$ & SSIM$\uparrow$ & LPIPS$\downarrow$\tabularnewline
\midrule
$\times$ & 1.0 & 20.310 & 0.872 & 0.1318\tabularnewline
$\checkmark$ & 1.0 & 20.352 & 0.873 & 0.1322\tabularnewline
$\checkmark$ & 1.5 & \textbf{20.528} & 0.874 & 0.1297\tabularnewline
$\checkmark$ & 2.5 & 20.508 & \textbf{0.875} & \textbf{0.1288}\tabularnewline
$\checkmark$ & 5.0 & 20.077 & 0.873 & 0.1310\tabularnewline
\bottomrule
\end{tabular}\tabularnewline
\end{table}

\begin{table}[htb]
\small
\centering
\caption{ Abaltions of CFG on MegaDepth~\citep{li2018megadepth} Ref-inpainting.\label{tab:ref_inpainting_CFG_ablation}}
\begin{tabular}{ccccc}
\toprule
Ref Views & CFG weight & PSNR$\uparrow$ & SSIM$\uparrow$ & LPIPS$\downarrow$\tabularnewline
\midrule
\multirow{4}{*}{1} & 1.0 & \textbf{21.502} & \textbf{0.840} & 0.1030\tabularnewline
 & 1.5 & 21.482 & \textbf{0.840} & 0.0955\tabularnewline
 & 2.0 & 21.195 & 0.837 & \textbf{0.0946}\tabularnewline
 & 2.5 & 20.761 & 0.832 & 0.0969\tabularnewline
 \midrule
\multirow{4}{*}{2} & 1.0 & \textbf{21.511} & \textbf{0.840} & 0.105\tabularnewline
 & 1.5 & 21.451 & \textbf{0.840} & 0.0977\tabularnewline
 & 2.0 & 21.092 & 0.836 & \textbf{0.0969}\tabularnewline
 & 2.5 & 20.614 & 0.830 & 0.0997\tabularnewline
 \midrule
 \multirow{4}{*}{3} & 1.0 & \textbf{21.771} & \textbf{0.844} & 0.0991\tabularnewline
 & 1.5 & 21.703 & \textbf{0.844} & 0.0912\tabularnewline
 & 2.0 & 21.356 & 0.840 & \textbf{0.0901}\tabularnewline
 & 2.5 & 20.855 & 0.834 & 0.0929\tabularnewline
 \midrule
 \multirow{4}{*}{4} & 1.0 & \textbf{22.334} & \textbf{0.851} & 0.0902\tabularnewline
 & 1.5 & 22.197 & 0.851 & \textbf{0.0836}\tabularnewline
 & 2.0 & 21.779 & 0.847 & 0.0839\tabularnewline
 & 2.5 & 21.125 & 0.841 & 0.0894\tabularnewline
\bottomrule
\end{tabular}\tabularnewline
\end{table}

\begin{figure}
\begin{center}
\includegraphics[width=0.85\linewidth]{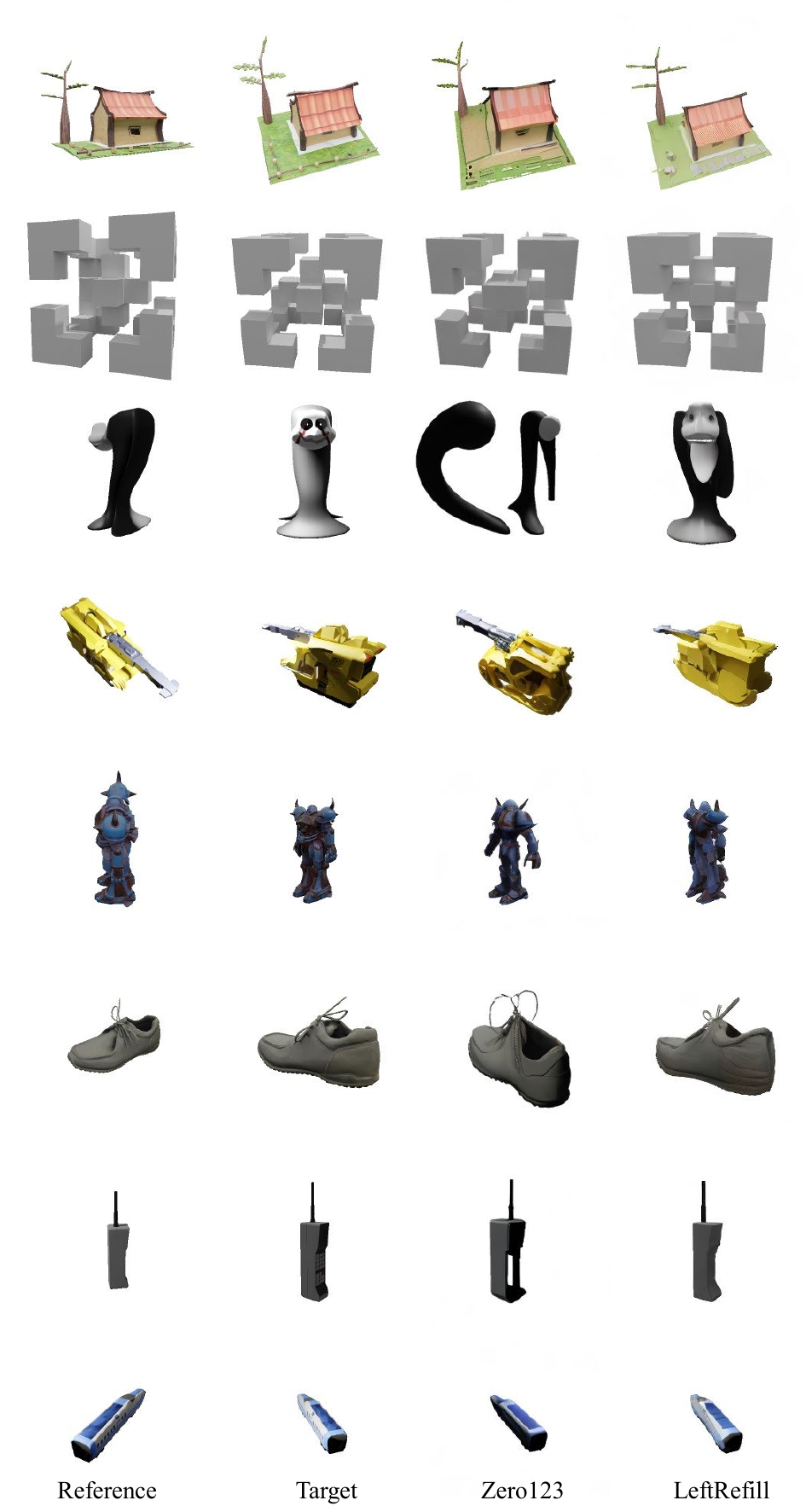}
\vspace{-0.1in}
\end{center}
   \caption{ NVS on Objaverse~\citep{deitke2022objaverse} from a single reference image.
   \label{fig:nvs_obj_supp}}
\vspace{-0.2in}
\end{figure}

\noindent\textbf{Matching-based Masks and Noise Coefficient.} 
On the left of \Cref{tab:reference_inpainting_config_ablation}, we find that the matching-based mask enjoys substantial improvement in the reference-guided inpainting. Besides, setting the noise coefficient $\eta=1$ achieves consistent improvements in our LeftRefill even sampled as the DDIM~\citep{song2020denoising}. So all LDMs are worked under $\eta=1$ without special illustrations.

\noindent\textbf{Prompt Initialization.} 
We tried three initialization ways for prompt tuning on the right of \Cref{tab:reference_inpainting_config_ablation}.
The random initialization performs worst. 
Both `token-wise' and `token-avgs' leverage text embeddings from a task-specific descriptive sentence listed as follows.
For the Ref-inpainting, the description is ``The whole image is split into two parts with the same size, they share the same scene/landmark captured with different viewpoints and times''.
For the NVS, the description is ``Left is the reference image, while the right one is the target image with a different viewpoint. The relative pose:''. Note that our encoded pose embedding is concatenated to the end of the task description embeddings.
`Token-wise' means repeating descriptive sentences until the prompt length, while each token is initialized for one prompt token. `Token-avgs' indicates that all prompt tokens are initialized with the average of the descriptive sentence. Meaningful initialization is useful for task-specific prompt tuning. 

\begin{figure*}
\begin{center}
\includegraphics[width=0.9\linewidth]{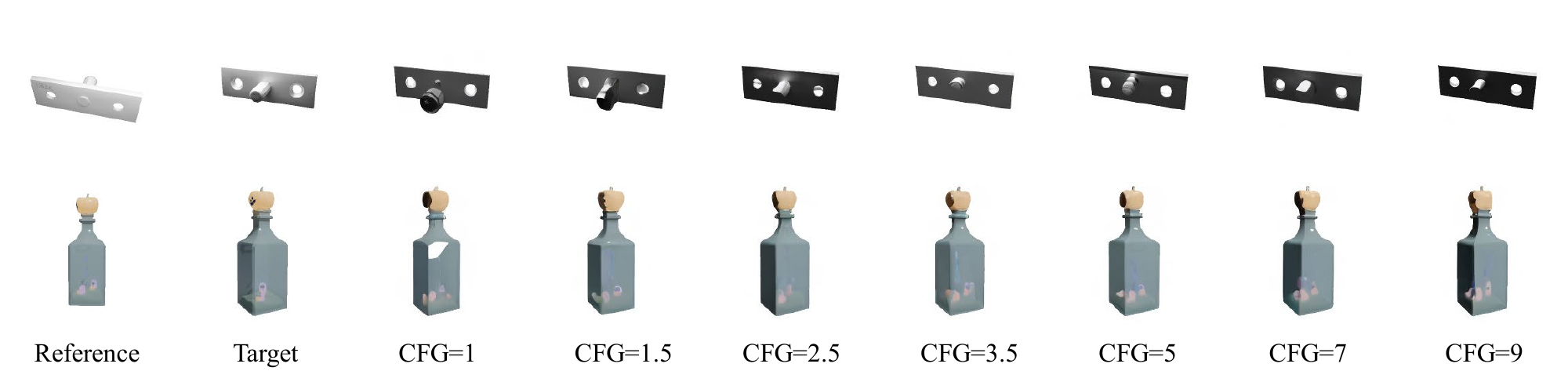}
\vspace{-0.15in}
\end{center}
   \caption{ NVS on Objaverse~\citep{deitke2022objaverse} with different CFG weights.
   \label{fig:nvs_cfg}}
\vspace{-0.15in}
\end{figure*}

\noindent\textbf{More Details about CFG.}
We remove the pose condition with 15\% to train the LeftRefill for NVS. Then the CFG coefficient 2.5 is used during the inference. As verified in \Cref{tab:cfg_nvs} and \Cref{fig:nvs_cfg}, appropriate CFG could improve the performance with better pose control and shape generation, while high CFG weights suffer from over-saturated issues.
Moreover, we find that CFG can also enhance the performance of Ref-inpainting even without training with prompts dropout as in \Cref{tab:ref_inpainting_CFG_ablation}. 
The LPIPS initially decreases but then increases as the CFG decreases from 2.5 to 1.0, while the PSNR and the SSIM keep increasing. We consider LPIPS as the most crucial metric, as it aligns with human perception.
Hence, when testing our model for Ref-inpainting, we opt to set CFG to 2.0. 
Furthermore, qualitative CFG results shown in the main paper also prove that 2.0 is a suitable trade-off between geometry and texture.

\begin{figure}
\begin{center}
\includegraphics[width=0.9\linewidth]{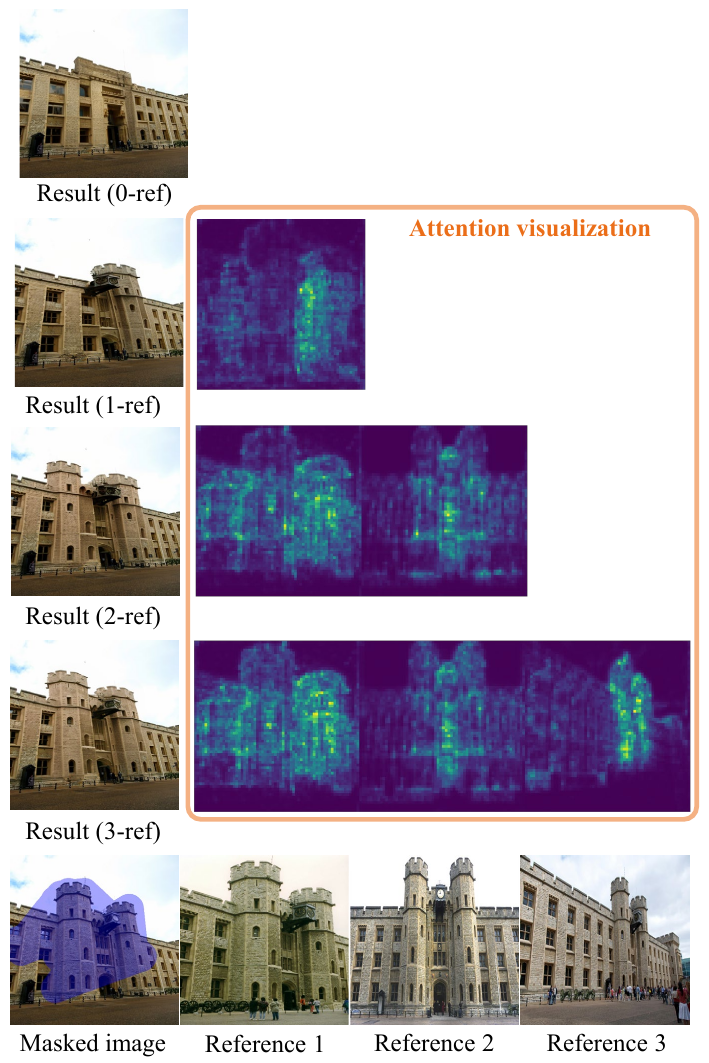}
\vspace{-0.15in}
\end{center}
   \caption{ Attention visualization (\Cref{alg:attention_score}) with increased reference views.
   \label{fig:attn_vis_increased_views}}
\vspace{-0.15in}
\end{figure}

\noindent\textbf{Attention Visualization with Increased References.}
We visualize the attention map for increased reference views under DDIM step 20 in \Cref{fig:attn_vis_increased_views}. More reference views help to rectify both inpainted results and attention maps. Note that we also show the result without any reference in \Cref{fig:attn_vis_increased_views}, which can be seen as vanilla inpainting. The prompt tuning fails to recover correct structures without reliable reference.

\subsection{Results of Ref-inpainting}
\label{sec:supp_ref_inpainting}

\begin{figure*}
\begin{center}
\includegraphics[width=0.95\linewidth]{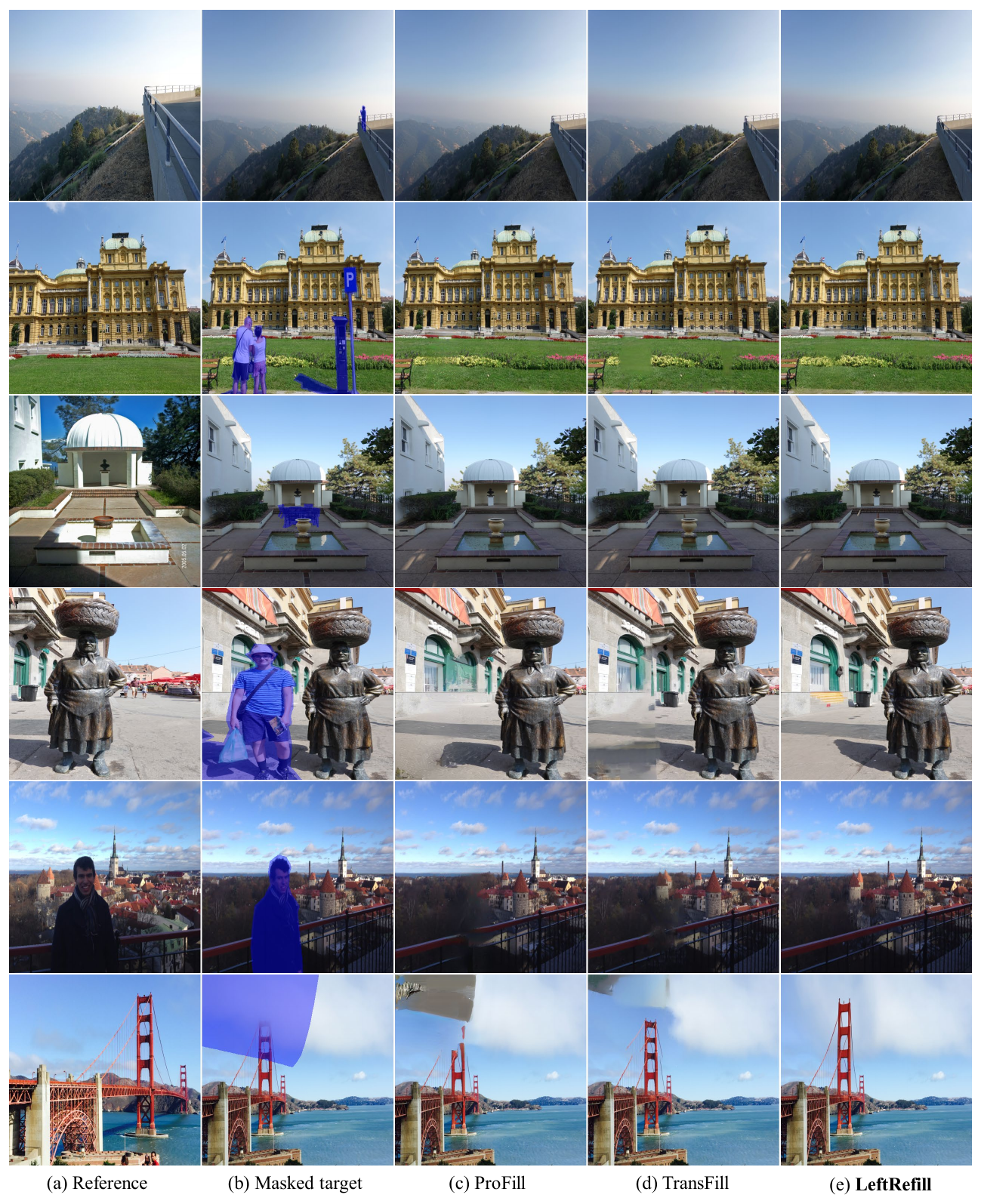}
\vspace{-0.1in}
\end{center}
   \caption{ Qualitative Ref-inpainting results compared with ProFill~\citep{zeng2020high}, TransFill~\citep{zhou2021transfill}, LeftRefill on the challenging real set provided by TransFill~\citep{zhou2021transfill}.
   \label{fig:real_set_qualitative}}
\vspace{-0.1in}
\end{figure*}

\begin{figure*}
\begin{center}
\includegraphics[width=1.0\linewidth]{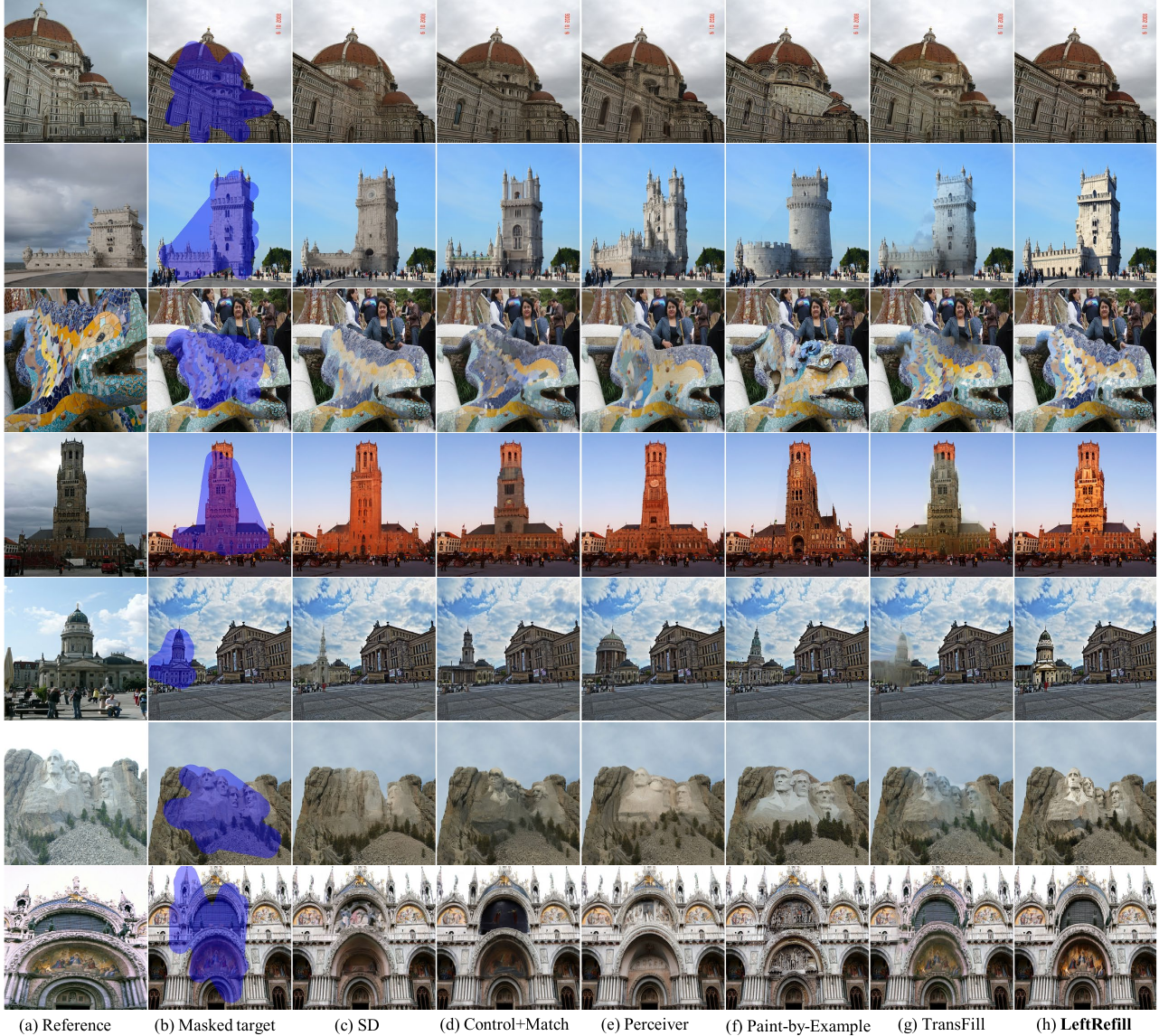}
\end{center}
   \caption{ Qualitative Ref-inpainting results on MegaDepth, which are compared among (c) SD~\citep{rombach2022high}, (d) ControlNet~\citep{zhang2023adding}+Matching~\citep{tang2022quadtree}, (e) Perceiver~\citep{jaegle2021perceiver} with ImageCLIP~\citep{radford2021learning}, (f) Paint-by-Example~\citep{yang2023paint}, (g) TransFill~\citep{zhou2021transfill}, and (I) our LeftRefill.
   Please zoom in for more details.
   \label{fig:ref_inpaint_supp}}
\end{figure*}

We provide more qualitative and quantitative results of Ref-inpainting\footnote{Since TransFill~\citep{zhou2021transfill} is not released, we send our images and masks to the authors and take their inpainted results for the evaluation.} in \Cref{fig:real_set_qualitative}, \Cref{fig:ref_inpaint_supp}, and \Cref{tab:reference_inpainting_realworld}.
Since most instances should be defined as object removal tasks without ground truth, quantitative metrics are for reference only. But LeftRefill still outperforms TransFill in FID and LPIPS with perceptually pleasant results. 
Moreover, as shown in \Cref{fig:real_set_qualitative}, LeftRefill enjoys good generalization in unseen or occluded regions, because it gets rid of the constrained geometric warping. 

\begin{figure}
\begin{center}
\includegraphics[width=0.95\linewidth]{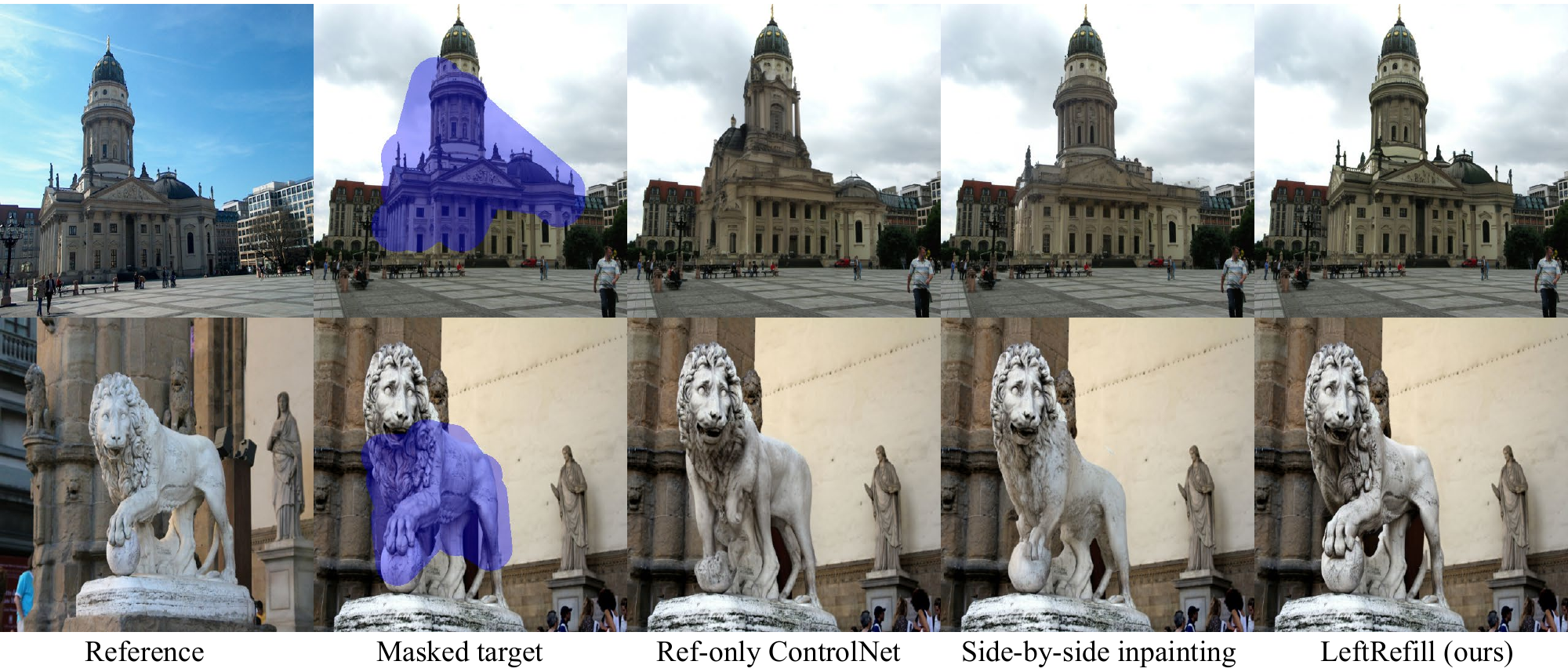}
\vspace{-0.15in}
\end{center}
   \caption{ Qualitative Ref-inpainting results compared to Ref-only ControlNet and side-by-side inpainting without prompt tuning.
   \label{fig:compare_ref_only}}
\vspace{-0.15in}
\end{figure}

\begin{table}
\small
\centering
\caption{ Quantitative Ref-inpainting results compared to Ref-only ControlNet and side-by-side inpainting without prompt tuning.\label{tab:compare_ref_only}}
\begin{tabular}{cccc}
\toprule
 & PSNR$\uparrow$ & SSIM$\uparrow$ & LPIPS$\downarrow$\tabularnewline
\midrule
Ref-only & 19.95 & 0.822 & 0.143\tabularnewline
Side-by-Side & 20.34 & 0.827 & 0.130\tabularnewline
LeftRefill & \textbf{20.93} & \textbf{0.836} & \textbf{0.096}\tabularnewline
\bottomrule
\end{tabular}\tabularnewline
\end{table}

\noindent\textbf{Ref-only ControlNet and Side-by-Side Inpainting.}
We further compare our LeftRefill to the popular Reference-only (Ref-only) ControlNet\footnote{https://github.com/Mikubill/sd-webui-controlnet/discussions/1236.} and side-by-side inpainting (without prompt tuning) as in \Cref{fig:compare_ref_only} and \Cref{tab:compare_ref_only}. However, they failed to address Ref-inpainting, retaining lower priority compared to other competitors in our main paper. Particularly, Ref-only ControlNet just limits attention fields, struggling to learn reasonable correlations. While side-by-side inpainting only stitches reference and target together without explicit instruction to control proper generation.

\begin{table}
\small 
\caption{ Ref-inpainting results on the real-world set~\citep{zhou2021transfill}.
\label{tab:reference_inpainting_realworld}}
\centering
\begin{tabular}{ccccc}
\toprule 
Method & PSNR$\uparrow$ & SSIM$\uparrow$ & FID$\downarrow$ & LPIPS$\downarrow$\tabularnewline
\midrule
ProFill~\citep{zeng2020high} & 25.550 & 0.944 & 71.758 & 0.0848\tabularnewline
TransFill~\citep{zhou2021transfill} & \textbf{26.052} & \textbf{0.945} & 62.493 & 0.0757\tabularnewline
\textbf{LeftRefill} & 25.733 & 0.942 & \textbf{61.276} & \textbf{0.0756}\tabularnewline
\bottomrule
\end{tabular}
\end{table}

\subsection{Results of NVS}
\label{sec:supp_nvs}

\begin{figure*}
\begin{center}
\includegraphics[width=0.8\linewidth]{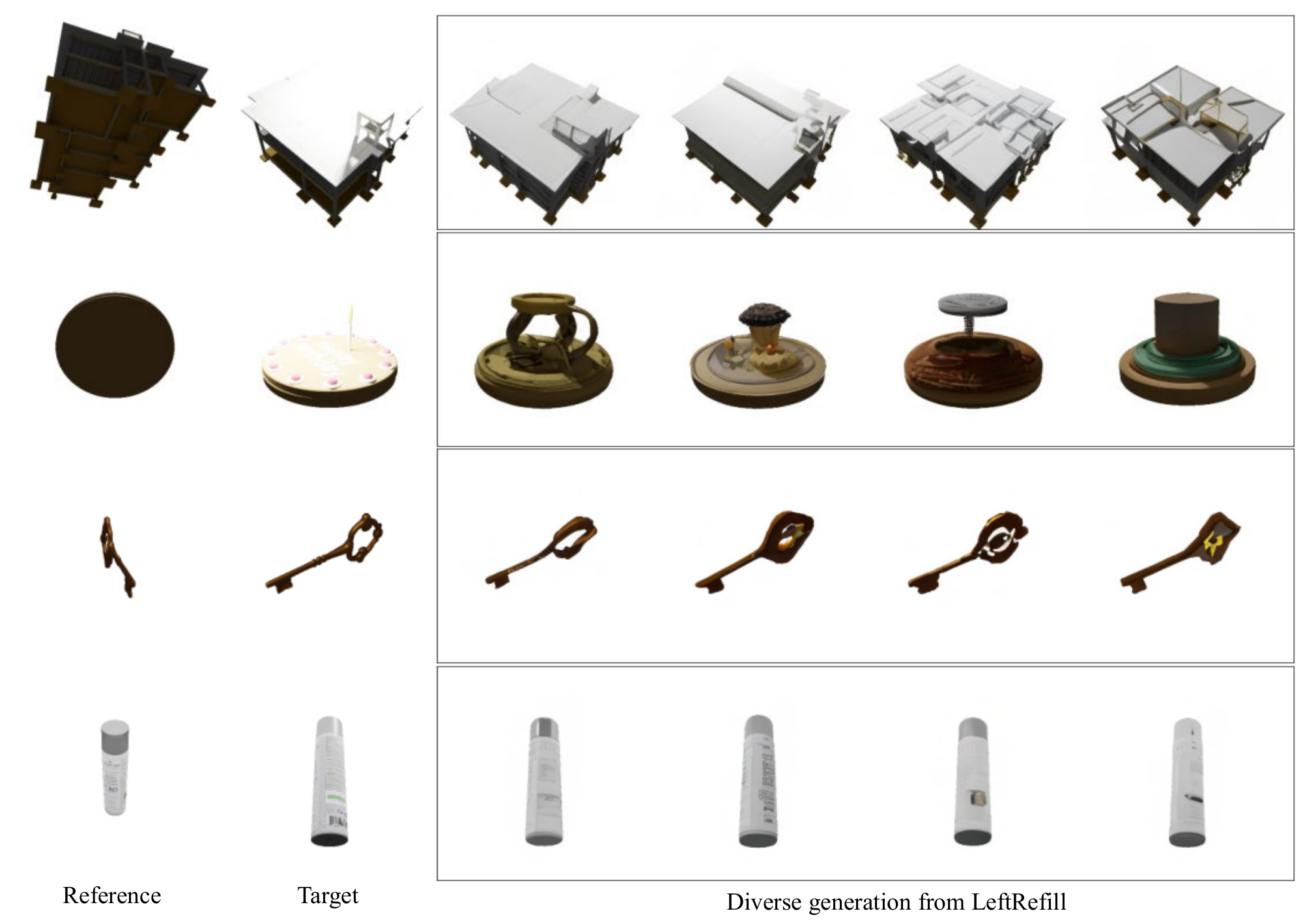}
\vspace{-0.1in}
\end{center}
   \caption{ Diversity of the NVS on Objaverse~\citep{deitke2022objaverse} from a single reference image without multi-view guidance.
   \label{fig:nvs_diversity}}
\vspace{-0.1in}
\end{figure*}

\begin{figure*}
\begin{center}
\includegraphics[width=1.0\linewidth]{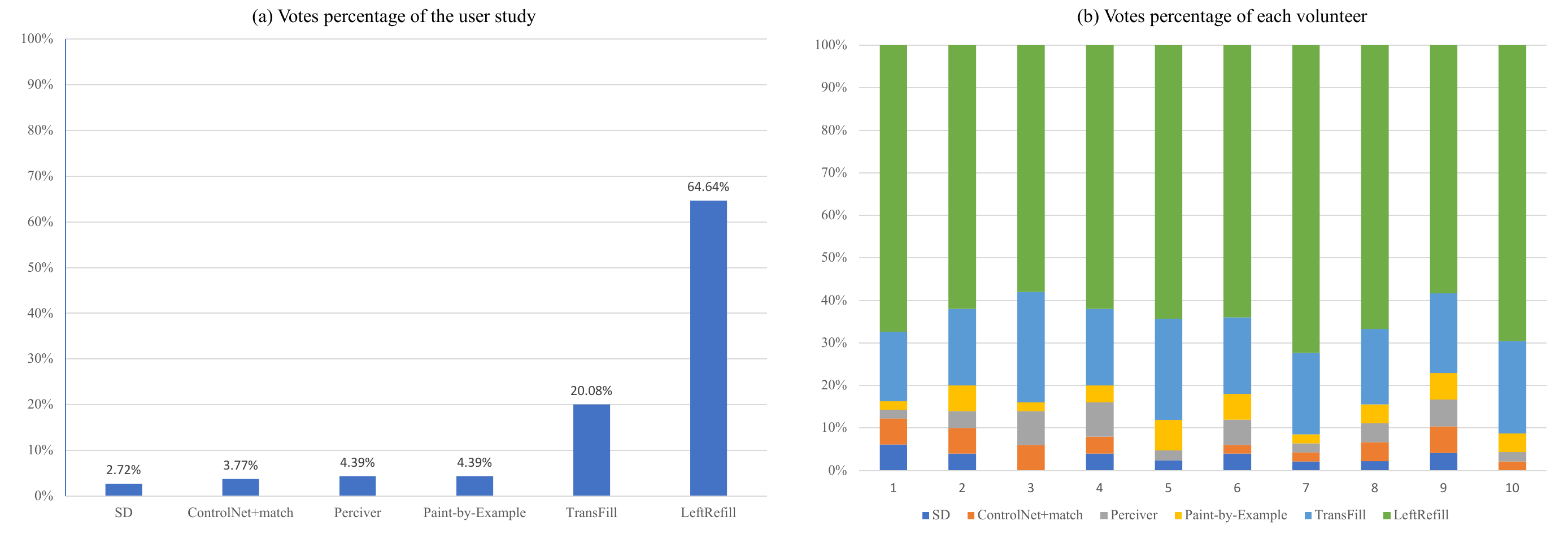}
\vspace{-0.1in}
\end{center}
   \caption{ The user study evaluation; (a) the overall voting percentage; (b) the votes of each volunteer.
   \label{fig:userstudy}}
\vspace{-0.1in}
\end{figure*}

\begin{figure*}
\begin{center}
\includegraphics[width=0.9\linewidth]{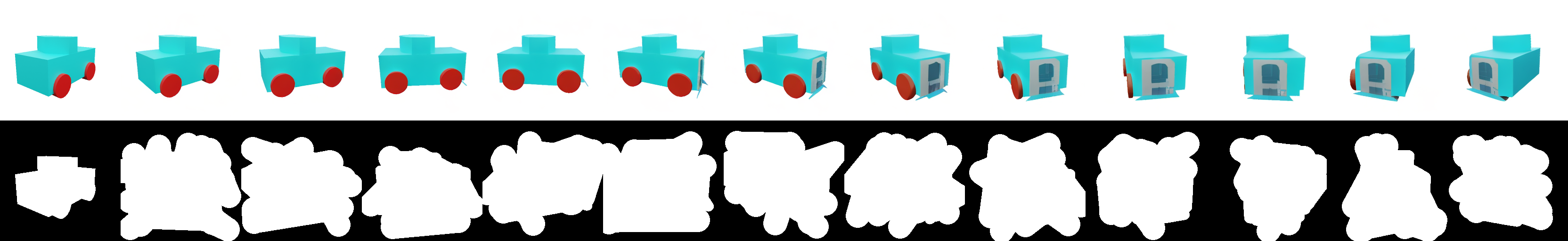}
\vspace{-0.1in}
\end{center}
   \caption{ The error accumulation occurred in AR generation. The degraded result is first generated in view 3.
   \label{fig:limitation}}
\vspace{-0.1in}
\end{figure*}

\begin{figure*}
\begin{center}
\includegraphics[width=0.75\linewidth]{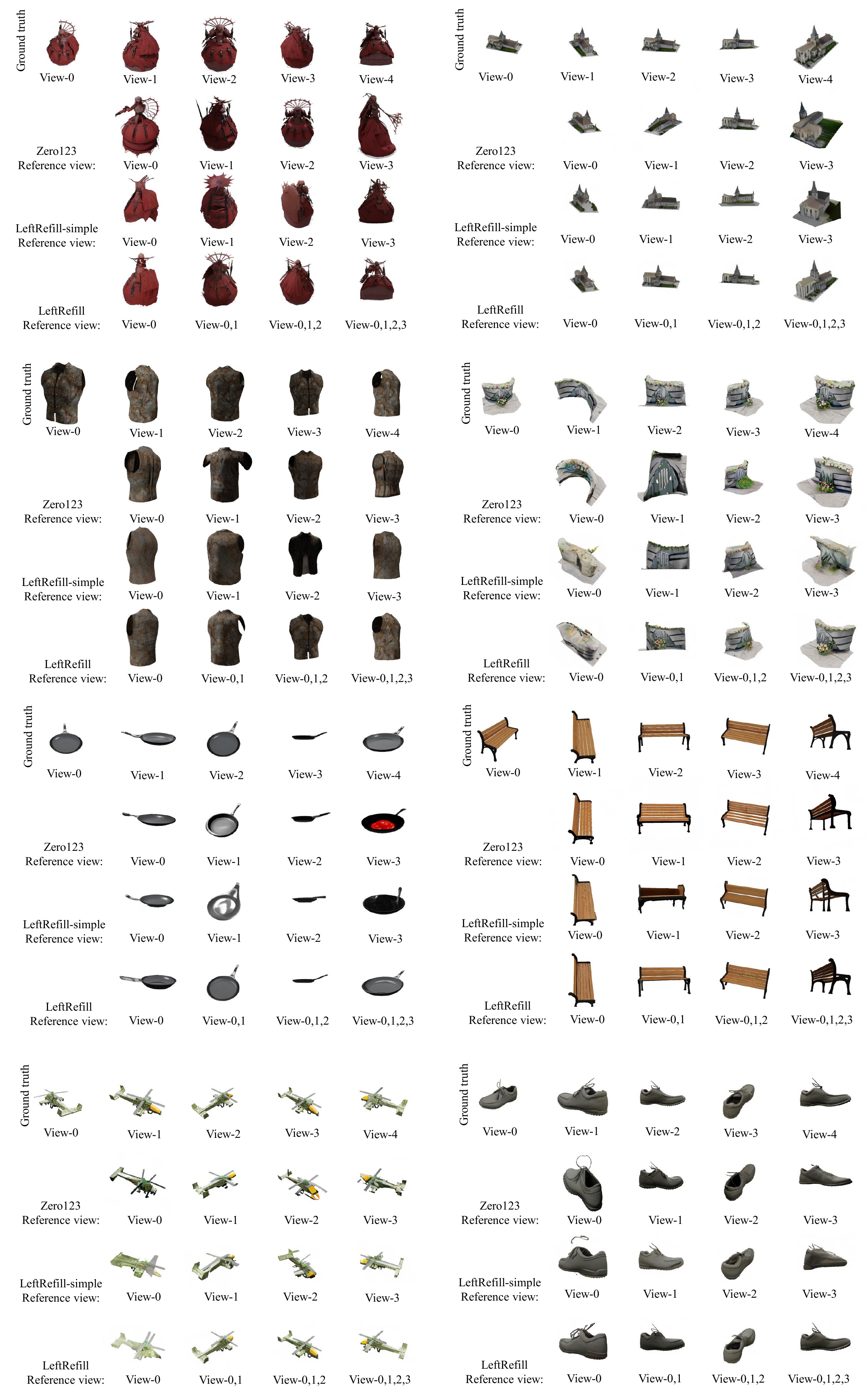}
\vspace{-0.15in}
\end{center}
   \caption{ Multi-view NVS results on Objaverse compared among the official Zero123~\citep{liu2023zero}, one-view based LeftRefill-simple, and multi-view based LeftRefill. Please zoom in for details.
   \label{fig:mv_nvs_obj_supp}}
\vspace{-0.15in}
\end{figure*}

Besides, we show some diverse NVS on Objaverse~\citep{deitke2022objaverse} in \Cref{fig:nvs_diversity}. 
Different random seeds are utilized to process the DDIM sampling. LeftRefill can achieve reasonable results with correct target poses.
More qualitative results are in \Cref{fig:nvs_obj_supp} and \Cref{fig:mv_nvs_obj_supp}.

\begin{table}
\small 
\centering
\caption{ The out-of-distribution comparison on Google Scanned Objects~\citep{downs2022google}.
\label{tab:nvs_gso_quantitative}}
\renewcommand\tabcolsep{2.3pt}
\begin{tabular}{cccccc}
\toprule 
Methods & Ref-View & PSNR$\uparrow$ & SSIM$\uparrow$ & LPIPS$\downarrow$ & CLIP$\uparrow$\tabularnewline
\midrule 
Zero123~\citep{liu2023zero} & 1 & 18.794 & 0.851 & 0.1132 & 0.7270\tabularnewline
LeftRefill & 1 & \textbf{21.039} & \textbf{0.883} & \textbf{0.0909} & \textbf{0.7693}\tabularnewline
\midrule 
LeftRefill & 2 & 22.090 & 0.893 & 0.0729 & 0.7925\tabularnewline
LeftRefill & 3 & 22.917 & \textbf{0.904} & 0.0595 & 0.8089\tabularnewline
LeftRefill & 4 & \textbf{23.169} & \textbf{0.904} & \textbf{0.0563} & \textbf{0.8185}\tabularnewline
\bottomrule 
\end{tabular}
\end{table}

\begin{figure*}
\begin{center}
\includegraphics[width=1.0\linewidth]{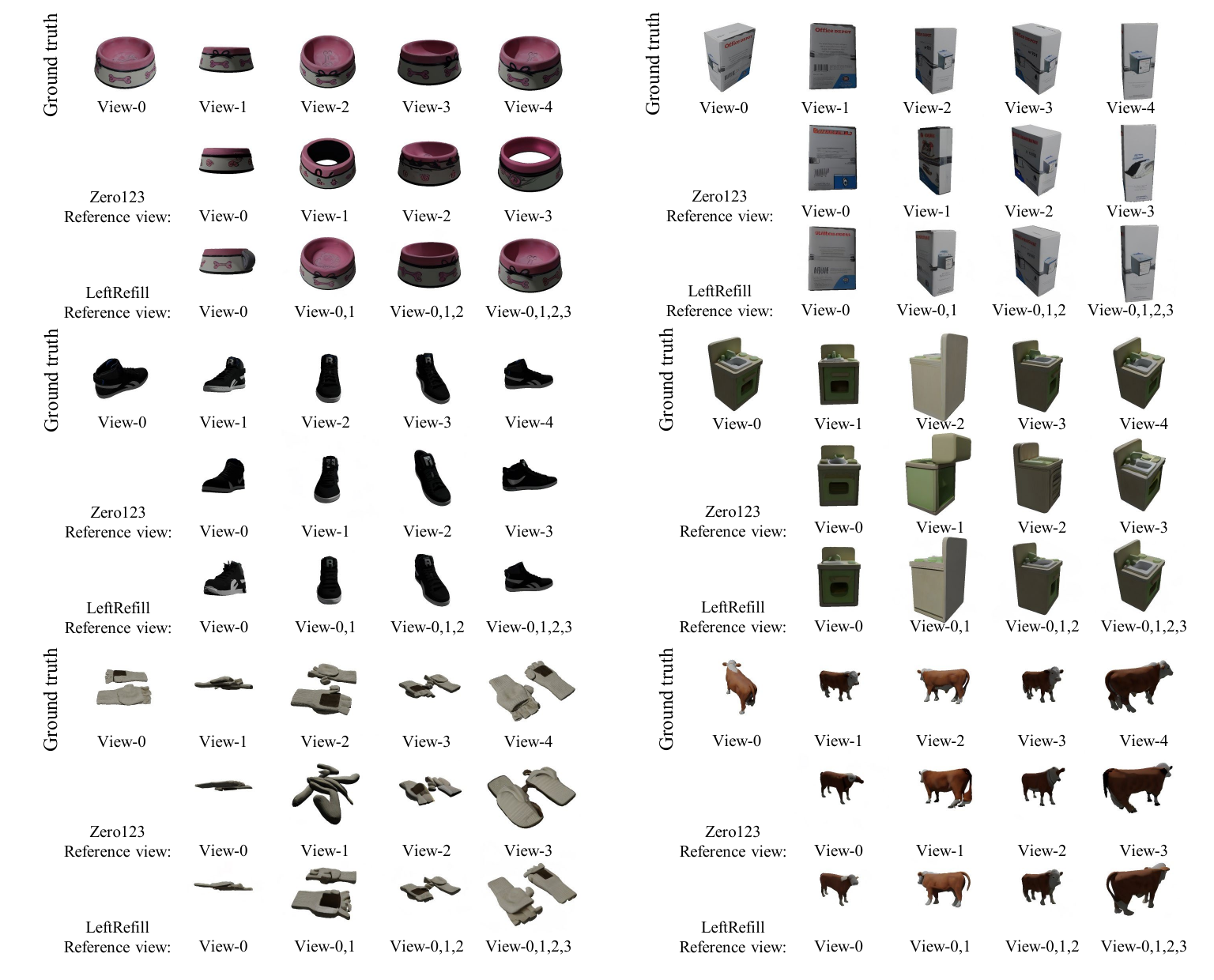}
\vspace{-0.1in}
\end{center}
   \caption{ Multi-view NVS results on Google Scanned Objects~\citep{downs2022google} compared with official Zero123~\citep{liu2023zero} and multi-view based LeftRefill. Please zoom in for details.
   \label{fig:nvs_gso_qualitative}}
\vspace{-0.1in}
\end{figure*}

\noindent\textbf{Comparison on Google Scanned Objects (GSO).}
We compare the proposed LeftRefill and zero123~\citep{liu2023zero} on the out-of-distribution GSO dataset~\citep{downs2022google} in \Cref{tab:nvs_gso_quantitative} and \Cref{fig:nvs_gso_qualitative}.
LeftRefill enjoys good zero-shot generalization, which outperforms zero123 with 1-view inputs. More reference views can further improve the quality of LeftRefill, benefiting from our multi-view-based NVS design and AR training.

\begin{table}
\small 
\caption{Inference speed of SD under 50 DDIM sampling steps.
\label{tab:infer_speed}}
\vspace{-0.1in}
\centering
\begin{tabular}{cccc}
\toprule
Input size & Sec/image & Input size & Sec/image\tabularnewline
\midrule
256$\times$256 & 2.9172 & 256$\times$512 & 2.9395\tabularnewline
\midrule
512$\times$512 & 3.0715 & 512$\times$1024 & 4.0205\tabularnewline
\bottomrule
\end{tabular}
\end{table}

\section{Inference Speed}

We provide the inference speed for different input resolutions in \Cref{tab:infer_speed}. 
All tests are based on one 32GB V100 GPU with 50 DDIM steps.
LeftRefill needs to stitch two images together, which would double the input size. But the inference time is not doubled as shown in \Cref{tab:infer_speed}.
Note that when the image size is smaller than 512, the difference in inference costs is not obvious.
Therefore, we think the proposed LeftRefill's inference cost is still acceptable in most real-world applications.

\section{User Study}
To evaluate the effectiveness of our LeftRefill in Ref-inpainting. We further test the user study as the human perceptual metric in \Cref{fig:userstudy}. 
Formally, 50 masked image pairs are randomly selected from our test set which are compared among SD~\citep{rombach2022high}, ControlNet~\citep{zhang2023adding}+match~\citep{tang2022quadtree}, Perciver~\citep{jaegle2021perceiver}, Paint-by-Example~\citep{yang2023paint}, TransFill~\citep{zhou2021transfill}, and LeftRefill. 
Although TransFill was not open-released, we thank TransFill's authors for kindly testing these samples for us. There are 10 volunteers who are not familiar with image generation attending this study.
Given masked target images and reference ones, we ask volunteers to vote for the best recovery from the 6 competitors mentioned above. 
The voting criterion should consider both the faithful recovery according to the reference and natural generations of color and texture. As shown in \Cref{fig:userstudy}, LeftRefill outperforms other competitors.

\section{Limitation}
\label{sec:limitation}

Although the proposed LeftRefill enjoys good performance and geometric consistency in multi-view NVS, it still suffers from the drawback of \emph{error accumulation} as shown in \Cref{fig:limitation}.
To eliminate this problem, we recommend providing a few more views (2,3,4) for more robust geometric priors.
Moreover, the extension to higher resolution and improved efficiency for pre-trained models with superior capacity (SDXL~\cite{podell2024sdxl}) can be regarded as interesting future work of LeftRefill.

\end{document}

%% file: preamble.tex
\usepackage[dvipsnames]{xcolor}


%% file: main.bbl
\begin{thebibliography}{71}
\providecommand{\natexlab}[1]{#1}
\providecommand{\url}[1]{\texttt{#1}}
\expandafter\ifx\csname urlstyle\endcsname\relax
  \providecommand{\doi}[1]{doi: #1}\else
  \providecommand{\doi}{doi: \begingroup \urlstyle{rm}\Url}\fi

\bibitem[Avrahami et~al.(2022)Avrahami, Lischinski, and
  Fried]{avrahami2022blended}
Omri Avrahami, Dani Lischinski, and Ohad Fried.
\newblock Blended diffusion for text-driven editing of natural images.
\newblock In \emph{Proceedings of the IEEE/CVF Conference on Computer Vision
  and Pattern Recognition}, pages 18208--18218, 2022.

\bibitem[Bar-Tal et~al.(2023)Bar-Tal, Yariv, Lipman, and
  Dekel]{bar2023multidiffusion}
Omer Bar-Tal, Lior Yariv, Yaron Lipman, and Tali Dekel.
\newblock Multidiffusion: Fusing diffusion paths for controlled image
  generation.
\newblock \emph{arXiv preprint arXiv:2302.08113}, 2, 2023.

\bibitem[Bertalmio et~al.(2000)Bertalmio, Sapiro, Caselles, and
  Ballester]{bertalmio2000image}
Marcelo Bertalmio, Guillermo Sapiro, Vincent Caselles, and Coloma Ballester.
\newblock Image inpainting.
\newblock In \emph{Proceedings of the 27th annual conference on Computer
  graphics and interactive techniques}, pages 417--424, 2000.

\bibitem[Chan et~al.(2022)Chan, Lin, Chan, Nagano, Pan, De~Mello, Gallo,
  Guibas, Tremblay, Khamis, et~al.]{chan2022efficient}
Eric~R Chan, Connor~Z Lin, Matthew~A Chan, Koki Nagano, Boxiao Pan, Shalini
  De~Mello, Orazio Gallo, Leonidas~J Guibas, Jonathan Tremblay, Sameh Khamis,
  et~al.
\newblock Efficient geometry-aware 3d generative adversarial networks.
\newblock In \emph{Proceedings of the IEEE/CVF Conference on Computer Vision
  and Pattern Recognition}, pages 16123--16133, 2022.

\bibitem[Chan et~al.(2023)Chan, Nagano, Chan, Bergman, Park, Levy, Aittala,
  De~Mello, Karras, and Wetzstein]{chan2023generative}
Eric~R Chan, Koki Nagano, Matthew~A Chan, Alexander~W Bergman, Jeong~Joon Park,
  Axel Levy, Miika Aittala, Shalini De~Mello, Tero Karras, and Gordon
  Wetzstein.
\newblock Generative novel view synthesis with 3d-aware diffusion models.
\newblock \emph{arXiv preprint arXiv:2304.02602}, 2023.

\bibitem[Chang et~al.(2023)Chang, Zhang, Barber, Maschinot, Lezama, Jiang,
  Yang, Murphy, Freeman, Rubinstein, et~al.]{chang2023muse}
Huiwen Chang, Han Zhang, Jarred Barber, AJ Maschinot, Jose Lezama, Lu Jiang,
  Ming-Hsuan Yang, Kevin Murphy, William~T Freeman, Michael Rubinstein, et~al.
\newblock Muse: Text-to-image generation via masked generative transformers.
\newblock \emph{arXiv preprint arXiv:2301.00704}, 2023.

\bibitem[Chen et~al.(2019)Chen, Ling, Gao, Smith, Lehtinen, Jacobson, and
  Fidler]{chen2019learning}
Wenzheng Chen, Huan Ling, Jun Gao, Edward Smith, Jaakko Lehtinen, Alec
  Jacobson, and Sanja Fidler.
\newblock Learning to predict 3d objects with an interpolation-based
  differentiable renderer.
\newblock \emph{Advances in neural information processing systems}, 32, 2019.

\bibitem[Couairon et~al.(2022)Couairon, Verbeek, Schwenk, and
  Cord]{couairon2022diffedit}
Guillaume Couairon, Jakob Verbeek, Holger Schwenk, and Matthieu Cord.
\newblock Diffedit: Diffusion-based semantic image editing with mask guidance.
\newblock \emph{arXiv preprint arXiv:2210.11427}, 2022.

\bibitem[Criminisi et~al.(2003)Criminisi, Perez, and
  Toyama]{criminisi2003object}
Antonio Criminisi, Patrick Perez, and Kentaro Toyama.
\newblock Object removal by exemplar-based inpainting.
\newblock In \emph{2003 IEEE Computer Society Conference on Computer Vision and
  Pattern Recognition, 2003. Proceedings.}, pages II--II. IEEE, 2003.

\bibitem[Deitke et~al.(2022)Deitke, Schwenk, Salvador, Weihs, Michel,
  VanderBilt, Schmidt, Ehsani, Kembhavi, and Farhadi]{deitke2022objaverse}
Matt Deitke, Dustin Schwenk, Jordi Salvador, Luca Weihs, Oscar Michel, Eli
  VanderBilt, Ludwig Schmidt, Kiana Ehsani, Aniruddha Kembhavi, and Ali
  Farhadi.
\newblock Objaverse: A universe of annotated 3d objects.
\newblock \emph{arXiv preprint arXiv:2212.08051}, 2022.

\bibitem[Dong et~al.(2022)Dong, Cao, and Fu]{dong2022incremental}
Qiaole Dong, Chenjie Cao, and Yanwei Fu.
\newblock Incremental transformer structure enhanced image inpainting with
  masking positional encoding.
\newblock In \emph{Proceedings of the IEEE/CVF Conference on Computer Vision
  and Pattern Recognition}, pages 11358--11368, 2022.

\bibitem[Downs et~al.(2022)Downs, Francis, Koenig, Kinman, Hickman, Reymann,
  McHugh, and Vanhoucke]{downs2022google}
Laura Downs, Anthony Francis, Nate Koenig, Brandon Kinman, Ryan Hickman, Krista
  Reymann, Thomas~B McHugh, and Vincent Vanhoucke.
\newblock Google scanned objects: A high-quality dataset of 3d scanned
  household items.
\newblock In \emph{2022 International Conference on Robotics and Automation
  (ICRA)}, pages 2553--2560. IEEE, 2022.

\bibitem[Esser et~al.(2021)Esser, Rombach, and Ommer]{esser2021taming}
Patrick Esser, Robin Rombach, and Bjorn Ommer.
\newblock Taming transformers for high-resolution image synthesis.
\newblock In \emph{Proceedings of the IEEE/CVF conference on computer vision
  and pattern recognition}, pages 12873--12883, 2021.

\bibitem[Fahim et~al.(2021)Fahim, Amin, and Zarif]{fahim2021single}
George Fahim, Khalid Amin, and Sameh Zarif.
\newblock Single-view 3d reconstruction: A survey of deep learning methods.
\newblock \emph{Computers \& Graphics}, 94:\penalty0 164--190, 2021.

\bibitem[Gal et~al.(2022)Gal, Alaluf, Atzmon, Patashnik, Bermano, Chechik, and
  Cohen-Or]{gal2022image}
Rinon Gal, Yuval Alaluf, Yuval Atzmon, Or Patashnik, Amit~H Bermano, Gal
  Chechik, and Daniel Cohen-Or.
\newblock An image is worth one word: Personalizing text-to-image generation
  using textual inversion.
\newblock \emph{arXiv preprint arXiv:2208.01618}, 2022.

\bibitem[Ge et~al.(2022)Ge, Huang, Xie, Lai, Song, Li, and Huang]{ge2022domain}
Chunjiang Ge, Rui Huang, Mixue Xie, Zihang Lai, Shiji Song, Shuang Li, and Gao
  Huang.
\newblock Domain adaptation via prompt learning.
\newblock \emph{arXiv preprint arXiv:2202.06687}, 2022.

\bibitem[Hays and Efros(2007)]{hays2007scene}
James Hays and Alexei~A Efros.
\newblock Scene completion using millions of photographs.
\newblock \emph{ACM Transactions on Graphics (ToG)}, 26\penalty0 (3):\penalty0
  4--es, 2007.

\bibitem[Hertz et~al.(2022)Hertz, Mokady, Tenenbaum, Aberman, Pritch, and
  Cohen-Or]{hertz2022prompt}
Amir Hertz, Ron Mokady, Jay Tenenbaum, Kfir Aberman, Yael Pritch, and Daniel
  Cohen-Or.
\newblock Prompt-to-prompt image editing with cross attention control.
\newblock \emph{arXiv preprint arXiv:2208.01626}, 2022.

\bibitem[Ho and Salimans(2022)]{ho2022classifier}
Jonathan Ho and Tim Salimans.
\newblock Classifier-free diffusion guidance.
\newblock \emph{arXiv preprint arXiv:2207.12598}, 2022.

\bibitem[Houlsby et~al.(2019)Houlsby, Giurgiu, Jastrzebski, Morrone,
  De~Laroussilhe, Gesmundo, Attariyan, and Gelly]{houlsby2019parameter}
Neil Houlsby, Andrei Giurgiu, Stanislaw Jastrzebski, Bruna Morrone, Quentin
  De~Laroussilhe, Andrea Gesmundo, Mona Attariyan, and Sylvain Gelly.
\newblock Parameter-efficient transfer learning for nlp.
\newblock In \emph{International Conference on Machine Learning}, pages
  2790--2799. PMLR, 2019.

\bibitem[Hu et~al.(2021)Hu, Shen, Wallis, Allen-Zhu, Li, Wang, Wang, and
  Chen]{hu2021lora}
Edward~J Hu, Yelong Shen, Phillip Wallis, Zeyuan Allen-Zhu, Yuanzhi Li, Shean
  Wang, Lu Wang, and Weizhu Chen.
\newblock Lora: Low-rank adaptation of large language models.
\newblock \emph{arXiv preprint arXiv:2106.09685}, 2021.

\bibitem[Jaegle et~al.(2021)Jaegle, Gimeno, Brock, Vinyals, Zisserman, and
  Carreira]{jaegle2021perceiver}
Andrew Jaegle, Felix Gimeno, Andy Brock, Oriol Vinyals, Andrew Zisserman, and
  Joao Carreira.
\newblock Perceiver: General perception with iterative attention.
\newblock In \emph{International conference on machine learning}, pages
  4651--4664. PMLR, 2021.

\bibitem[Jia et~al.(2022)Jia, Tang, Chen, Cardie, Belongie, Hariharan, and
  Lim]{jia2022visual}
Menglin Jia, Luming Tang, Bor-Chun Chen, Claire Cardie, Serge Belongie, Bharath
  Hariharan, and Ser-Nam Lim.
\newblock Visual prompt tuning.
\newblock In \emph{Computer Vision--ECCV 2022: 17th European Conference, Tel
  Aviv, Israel, October 23--27, 2022, Proceedings, Part XXXIII}, pages
  709--727. Springer, 2022.

\bibitem[Lester et~al.(2021)Lester, Al-Rfou, and Constant]{lester2021power}
Brian Lester, Rami Al-Rfou, and Noah Constant.
\newblock The power of scale for parameter-efficient prompt tuning.
\newblock \emph{arXiv preprint arXiv:2104.08691}, 2021.

\bibitem[Li et~al.(2022)Li, Lin, Zhou, Qi, Wang, and Jia]{li2022mat}
Wenbo Li, Zhe Lin, Kun Zhou, Lu Qi, Yi Wang, and Jiaya Jia.
\newblock Mat: Mask-aware transformer for large hole image inpainting.
\newblock In \emph{Proceedings of the IEEE/CVF conference on computer vision
  and pattern recognition}, pages 10758--10768, 2022.

\bibitem[Li et~al.(2023)Li, Liu, Wu, Mu, Yang, Gao, Li, and Lee]{li2023gligen}
Yuheng Li, Haotian Liu, Qingyang Wu, Fangzhou Mu, Jianwei Yang, Jianfeng Gao,
  Chunyuan Li, and Yong~Jae Lee.
\newblock Gligen: Open-set grounded text-to-image generation.
\newblock \emph{arXiv preprint arXiv:2301.07093}, 2023.

\bibitem[Li and Snavely(2018)]{li2018megadepth}
Zhengqi Li and Noah Snavely.
\newblock Megadepth: Learning single-view depth prediction from internet
  photos.
\newblock In \emph{Proceedings of the IEEE conference on computer vision and
  pattern recognition}, pages 2041--2050, 2018.

\bibitem[Liao et~al.(2023)Liao, Shi, Cao, Zhang, Tian, and
  Yan]{liao2023rethinking}
Ning Liao, Bowen Shi, Min Cao, Xiaopeng Zhang, Qi Tian, and Junchi Yan.
\newblock Rethinking visual prompt learning as masked visual token modeling.
\newblock \emph{arXiv preprint arXiv:2303.04998}, 2023.

\bibitem[Liu et~al.(2023{\natexlab{a}})Liu, Yuan, Fu, Jiang, Hayashi, and
  Neubig]{liu2023pre}
Pengfei Liu, Weizhe Yuan, Jinlan Fu, Zhengbao Jiang, Hiroaki Hayashi, and
  Graham Neubig.
\newblock Pre-train, prompt, and predict: A systematic survey of prompting
  methods in natural language processing.
\newblock \emph{ACM Computing Surveys}, 55\penalty0 (9):\penalty0 1--35,
  2023{\natexlab{a}}.

\bibitem[Liu et~al.(2023{\natexlab{b}})Liu, Wu, Van~Hoorick, Tokmakov,
  Zakharov, and Vondrick]{liu2023zero}
Ruoshi Liu, Rundi Wu, Basile Van~Hoorick, Pavel Tokmakov, Sergey Zakharov, and
  Carl Vondrick.
\newblock Zero-1-to-3: Zero-shot one image to 3d object.
\newblock In \emph{Proceedings of the IEEE/CVF international conference on
  computer vision}, 2023{\natexlab{b}}.

\bibitem[Liu et~al.(2019)Liu, Li, Chen, and Li]{liu2019soft}
Shichen Liu, Tianye Li, Weikai Chen, and Hao Li.
\newblock Soft rasterizer: A differentiable renderer for image-based 3d
  reasoning.
\newblock In \emph{Proceedings of the IEEE/CVF International Conference on
  Computer Vision}, pages 7708--7717, 2019.

\bibitem[Liu et~al.(2021{\natexlab{a}})Liu, Ji, Fu, Tam, Du, Yang, and
  Tang]{liu2021p}
Xiao Liu, Kaixuan Ji, Yicheng Fu, Weng~Lam Tam, Zhengxiao Du, Zhilin Yang, and
  Jie Tang.
\newblock P-tuning v2: Prompt tuning can be comparable to fine-tuning
  universally across scales and tasks.
\newblock \emph{arXiv preprint arXiv:2110.07602}, 2021{\natexlab{a}}.

\bibitem[Liu et~al.(2021{\natexlab{b}})Liu, Zheng, Du, Ding, Qian, Yang, and
  Tang]{liu2021gpt}
Xiao Liu, Yanan Zheng, Zhengxiao Du, Ming Ding, Yujie Qian, Zhilin Yang, and
  Jie Tang.
\newblock Gpt understands, too.
\newblock \emph{arXiv preprint arXiv:2103.10385}, 2021{\natexlab{b}}.

\bibitem[Ma et~al.(2023)Ma, Yang, Wang, Fu, and Liu]{ma2023unified}
Yiyang Ma, Huan Yang, Wenjing Wang, Jianlong Fu, and Jiaying Liu.
\newblock Unified multi-modal latent diffusion for joint subject and text
  conditional image generation.
\newblock \emph{arXiv preprint arXiv:2303.09319}, 2023.

\bibitem[Mokady et~al.(2022)Mokady, Hertz, Aberman, Pritch, and
  Cohen-Or]{mokady2022null}
Ron Mokady, Amir Hertz, Kfir Aberman, Yael Pritch, and Daniel Cohen-Or.
\newblock Null-text inversion for editing real images using guided diffusion
  models.
\newblock \emph{arXiv preprint arXiv:2211.09794}, 2022.

\bibitem[Mou et~al.(2023)Mou, Wang, Xie, Zhang, Qi, Shan, and Qie]{mou2023t2i}
Chong Mou, Xintao Wang, Liangbin Xie, Jian Zhang, Zhongang Qi, Ying Shan, and
  Xiaohu Qie.
\newblock T2i-adapter: Learning adapters to dig out more controllable ability
  for text-to-image diffusion models.
\newblock \emph{arXiv preprint arXiv:2302.08453}, 2023.

\bibitem[Nichol et~al.(2021)Nichol, Dhariwal, Ramesh, Shyam, Mishkin, McGrew,
  Sutskever, and Chen]{nichol2021glide}
Alex Nichol, Prafulla Dhariwal, Aditya Ramesh, Pranav Shyam, Pamela Mishkin,
  Bob McGrew, Ilya Sutskever, and Mark Chen.
\newblock Glide: Towards photorealistic image generation and editing with
  text-guided diffusion models.
\newblock \emph{arXiv preprint arXiv:2112.10741}, 2021.

\bibitem[Niemeyer and Geiger(2021)]{niemeyer2021giraffe}
Michael Niemeyer and Andreas Geiger.
\newblock Giraffe: Representing scenes as compositional generative neural
  feature fields.
\newblock In \emph{Proceedings of the IEEE/CVF Conference on Computer Vision
  and Pattern Recognition}, pages 11453--11464, 2021.

\bibitem[Niklaus et~al.(2019)Niklaus, Mai, Yang, and Liu]{niklaus20193d}
Simon Niklaus, Long Mai, Jimei Yang, and Feng Liu.
\newblock 3d ken burns effect from a single image.
\newblock \emph{ACM Transactions on Graphics (ToG)}, 38\penalty0 (6):\penalty0
  1--15, 2019.

\bibitem[Oh et~al.(2019)Oh, Lee, Lee, and Kim]{oh2019onion}
Seoung~Wug Oh, Sungho Lee, Joon-Young Lee, and Seon~Joo Kim.
\newblock Onion-peel networks for deep video completion.
\newblock In \emph{proceedings of the IEEE/cvf international conference on
  computer vision}, pages 4403--4412, 2019.

\bibitem[Podell et~al.(2023)Podell, English, Lacey, Blattmann, Dockhorn,
  M{\"u}ller, Penna, and Rombach]{podell2023sdxl}
Dustin Podell, Zion English, Kyle Lacey, Andreas Blattmann, Tim Dockhorn, Jonas
  M{\"u}ller, Joe Penna, and Robin Rombach.
\newblock Sdxl: Improving latent diffusion models for high-resolution image
  synthesis.
\newblock \emph{arXiv preprint arXiv:2307.01952}, 2023.

\bibitem[Podell et~al.(2024)Podell, English, Lacey, Blattmann, Dockhorn,
  M{\"u}ller, Penna, and Rombach]{podell2024sdxl}
Dustin Podell, Zion English, Kyle Lacey, Andreas Blattmann, Tim Dockhorn, Jonas
  M{\"u}ller, Joe Penna, and Robin Rombach.
\newblock {SDXL}: Improving latent diffusion models for high-resolution image
  synthesis.
\newblock In \emph{International Conference on Learning Representations
  (ICLR)}, 2024.

\bibitem[Poole et~al.(2023)Poole, Jain, Barron, and
  Mildenhall]{poole2023dreamfusion}
Ben Poole, Ajay Jain, Jonathan~T Barron, and Ben Mildenhall.
\newblock Dreamfusion: Text-to-3d using 2d diffusion.
\newblock In \emph{International Conference on Learning Representations
  (ICLR)}, 2023.

\bibitem[Qin et~al.(2020)Qin, Zhang, Huang, Dehghan, Zaiane, and
  Jagersand]{qin2020u2}
Xuebin Qin, Zichen Zhang, Chenyang Huang, Masood Dehghan, Osmar~R Zaiane, and
  Martin Jagersand.
\newblock U2-net: Going deeper with nested u-structure for salient object
  detection.
\newblock \emph{Pattern recognition}, 106:\penalty0 107404, 2020.

\bibitem[Radford et~al.(2021)Radford, Kim, Hallacy, Ramesh, Goh, Agarwal,
  Sastry, Askell, Mishkin, Clark, et~al.]{radford2021learning}
Alec Radford, Jong~Wook Kim, Chris Hallacy, Aditya Ramesh, Gabriel Goh,
  Sandhini Agarwal, Girish Sastry, Amanda Askell, Pamela Mishkin, Jack Clark,
  et~al.
\newblock Learning transferable visual models from natural language
  supervision.
\newblock In \emph{International conference on machine learning}, pages
  8748--8763. PMLR, 2021.

\bibitem[Ramesh et~al.(2022)Ramesh, Dhariwal, Nichol, Chu, and
  Chen]{ramesh2022hierarchical}
Aditya Ramesh, Prafulla Dhariwal, Alex Nichol, Casey Chu, and Mark Chen.
\newblock Hierarchical text-conditional image generation with clip latents.
\newblock \emph{arXiv preprint arXiv:2204.06125}, 2022.

\bibitem[Rombach et~al.(2021)Rombach, Esser, and Ommer]{rombach2021geometry}
Robin Rombach, Patrick Esser, and Bj{\"o}rn Ommer.
\newblock Geometry-free view synthesis: Transformers and no 3d priors.
\newblock In \emph{Proceedings of the IEEE/CVF International Conference on
  Computer Vision}, pages 14356--14366, 2021.

\bibitem[Rombach et~al.(2022)Rombach, Blattmann, Lorenz, Esser, and
  Ommer]{rombach2022high}
Robin Rombach, Andreas Blattmann, Dominik Lorenz, Patrick Esser, and Bj{\"o}rn
  Ommer.
\newblock High-resolution image synthesis with latent diffusion models.
\newblock In \emph{Proceedings of the IEEE/CVF Conference on Computer Vision
  and Pattern Recognition}, pages 10684--10695, 2022.

\bibitem[Ruiz et~al.(2022)Ruiz, Li, Jampani, Pritch, Rubinstein, and
  Aberman]{ruiz2022dreambooth}
Nataniel Ruiz, Yuanzhen Li, Varun Jampani, Yael Pritch, Michael Rubinstein, and
  Kfir Aberman.
\newblock Dreambooth: Fine tuning text-to-image diffusion models for
  subject-driven generation.
\newblock \emph{arXiv preprint arXiv:2208.12242}, 2022.

\bibitem[Saharia et~al.(2022)Saharia, Chan, Saxena, Li, Whang, Denton,
  Ghasemipour, Gontijo~Lopes, Karagol~Ayan, Salimans,
  et~al.]{saharia2022photorealistic}
Chitwan Saharia, William Chan, Saurabh Saxena, Lala Li, Jay Whang, Emily~L
  Denton, Kamyar Ghasemipour, Raphael Gontijo~Lopes, Burcu Karagol~Ayan, Tim
  Salimans, et~al.
\newblock Photorealistic text-to-image diffusion models with deep language
  understanding.
\newblock \emph{Advances in Neural Information Processing Systems},
  35:\penalty0 36479--36494, 2022.

\bibitem[Salimans et~al.(2017)Salimans, Karpathy, Chen, and
  Kingma]{salimans2017pixelcnn++}
Tim Salimans, Andrej Karpathy, Xi Chen, and Diederik~P Kingma.
\newblock Pixelcnn++: Improving the pixelcnn with discretized logistic mixture
  likelihood and other modifications.
\newblock In \emph{International Conference on Learning Representations
  (ICLR)}, 2017.

\bibitem[Schops et~al.(2017)Schops, Schonberger, Galliani, Sattler, Schindler,
  Pollefeys, and Geiger]{schops2017multi}
Thomas Schops, Johannes~L Schonberger, Silvano Galliani, Torsten Sattler,
  Konrad Schindler, Marc Pollefeys, and Andreas Geiger.
\newblock A multi-view stereo benchmark with high-resolution images and
  multi-camera videos.
\newblock In \emph{Proceedings of the IEEE Conference on Computer Vision and
  Pattern Recognition}, pages 3260--3269, 2017.

\bibitem[Schwarz et~al.(2020)Schwarz, Liao, Niemeyer, and
  Geiger]{schwarz2020graf}
Katja Schwarz, Yiyi Liao, Michael Niemeyer, and Andreas Geiger.
\newblock Graf: Generative radiance fields for 3d-aware image synthesis.
\newblock \emph{Advances in Neural Information Processing Systems},
  33:\penalty0 20154--20166, 2020.

\bibitem[Shih et~al.(2020)Shih, Su, Kopf, and Huang]{shih20203d}
Meng-Li Shih, Shih-Yang Su, Johannes Kopf, and Jia-Bin Huang.
\newblock 3d photography using context-aware layered depth inpainting.
\newblock In \emph{Proceedings of the IEEE/CVF Conference on Computer Vision
  and Pattern Recognition}, pages 8028--8038, 2020.

\bibitem[Sohn et~al.(2022)Sohn, Hao, Lezama, Polania, Chang, Zhang, Essa, and
  Jiang]{sohn2022visual}
Kihyuk Sohn, Yuan Hao, Jos{\'e} Lezama, Luisa Polania, Huiwen Chang, Han Zhang,
  Irfan Essa, and Lu Jiang.
\newblock Visual prompt tuning for generative transfer learning.
\newblock \emph{arXiv preprint arXiv:2210.00990}, 2022.

\bibitem[Song et~al.(2020)Song, Meng, and Ermon]{song2020denoising}
Jiaming Song, Chenlin Meng, and Stefano Ermon.
\newblock Denoising diffusion implicit models.
\newblock \emph{arXiv preprint arXiv:2010.02502}, 2020.

\bibitem[Suvorov et~al.(2022)Suvorov, Logacheva, Mashikhin, Remizova, Ashukha,
  Silvestrov, Kong, Goka, Park, and Lempitsky]{suvorov2022resolution}
Roman Suvorov, Elizaveta Logacheva, Anton Mashikhin, Anastasia Remizova,
  Arsenii Ashukha, Aleksei Silvestrov, Naejin Kong, Harshith Goka, Kiwoong
  Park, and Victor Lempitsky.
\newblock Resolution-robust large mask inpainting with fourier convolutions.
\newblock In \emph{Proceedings of the IEEE/CVF winter conference on
  applications of computer vision}, pages 2149--2159, 2022.

\bibitem[Tang et~al.(2022)Tang, Zhang, Zhu, and Tan]{tang2022quadtree}
Shitao Tang, Jiahui Zhang, Siyu Zhu, and Ping Tan.
\newblock Quadtree attention for vision transformers.
\newblock \emph{arXiv preprint arXiv:2201.02767}, 2022.

\bibitem[Van~den Oord et~al.(2016)Van~den Oord, Kalchbrenner, Espeholt,
  Vinyals, Graves, et~al.]{van2016conditional}
Aaron Van~den Oord, Nal Kalchbrenner, Lasse Espeholt, Oriol Vinyals, Alex
  Graves, et~al.
\newblock Conditional image generation with pixelcnn decoders.
\newblock \emph{Advances in neural information processing systems}, 29, 2016.

\bibitem[Vaswani et~al.(2017)Vaswani, Shazeer, Parmar, Uszkoreit, Jones, Gomez,
  Kaiser, and Polosukhin]{vaswani2017attention}
Ashish Vaswani, Noam Shazeer, Niki Parmar, Jakob Uszkoreit, Llion Jones,
  Aidan~N Gomez, {\L}ukasz Kaiser, and Illia Polosukhin.
\newblock Attention is all you need.
\newblock \emph{Advances in neural information processing systems}, 30, 2017.

\bibitem[Voynov et~al.(2022)Voynov, Aberman, and Cohen-Or]{voynov2022sketch}
Andrey Voynov, Kfir Aberman, and Daniel Cohen-Or.
\newblock Sketch-guided text-to-image diffusion models.
\newblock \emph{arXiv preprint arXiv:2211.13752}, 2022.

\bibitem[Wang et~al.(2018)Wang, Zhang, Li, Fu, Liu, and
  Jiang]{wang2018pixel2mesh}
Nanyang Wang, Yinda Zhang, Zhuwen Li, Yanwei Fu, Wei Liu, and Yu-Gang Jiang.
\newblock Pixel2mesh: Generating 3d mesh models from single rgb images.
\newblock In \emph{Proceedings of the European conference on computer vision
  (ECCV)}, pages 52--67, 2018.

\bibitem[Wu et~al.(2017)Wu, Wang, Xue, Sun, Freeman, and
  Tenenbaum]{wu2017marrnet}
Jiajun Wu, Yifan Wang, Tianfan Xue, Xingyuan Sun, Bill Freeman, and Josh
  Tenenbaum.
\newblock Marrnet: 3d shape reconstruction via 2.5 d sketches.
\newblock In \emph{Advances in neural information processing systems}, 2017.

\bibitem[Xu et~al.(2019)Xu, Wang, Ceylan, Mech, and Neumann]{xu2019disn}
Qiangeng Xu, Weiyue Wang, Duygu Ceylan, Radomir Mech, and Ulrich Neumann.
\newblock Disn: Deep implicit surface network for high-quality single-view 3d
  reconstruction.
\newblock In \emph{Advances in neural information processing systems}, 2019.

\bibitem[Yang et~al.(2023)Yang, Gu, Zhang, Zhang, Chen, Sun, Chen, and
  Wen]{yang2023paint}
Binxin Yang, Shuyang Gu, Bo Zhang, Ting Zhang, Xuejin Chen, Xiaoyan Sun, Dong
  Chen, and Fang Wen.
\newblock Paint by example: Exemplar-based image editing with diffusion models.
\newblock In \emph{Proceedings of the IEEE/CVF Conference on Computer Vision
  and Pattern Recognition}, pages 18381--18391, 2023.

\bibitem[Zeng et~al.(2020)Zeng, Lin, Yang, Zhang, Shechtman, and
  Lu]{zeng2020high}
Yu Zeng, Zhe Lin, Jimei Yang, Jianming Zhang, Eli Shechtman, and Huchuan Lu.
\newblock High-resolution image inpainting with iterative confidence feedback
  and guided upsampling.
\newblock In \emph{Computer Vision--ECCV 2020: 16th European Conference,
  Glasgow, UK, August 23--28, 2020, Proceedings, Part XIX 16}, pages 1--17.
  Springer, 2020.

\bibitem[Zhang and Agrawala(2023)]{zhang2023adding}
Lvmin Zhang and Maneesh Agrawala.
\newblock Adding conditional control to text-to-image diffusion models.
\newblock In \emph{Proceedings of the IEEE/CVF international conference on
  computer vision}, 2023.

\bibitem[Zhao et~al.(2022{\natexlab{a}})Zhao, Zhao, Ma, Zhang, and
  Zeng]{zhao20223dfill}
Liang Zhao, Xinyuan Zhao, Hailong Ma, Xinyu Zhang, and Long Zeng.
\newblock 3dfill: Reference-guided image inpainting by self-supervised 3d image
  alignment.
\newblock \emph{arXiv preprint arXiv:2211.04831}, 2022{\natexlab{a}}.

\bibitem[Zhao et~al.(2021)Zhao, Cui, Sheng, Dong, Liang, Chang, and
  Xu]{zhao2021large}
Shengyu Zhao, Jonathan Cui, Yilun Sheng, Yue Dong, Xiao Liang, Eric~I Chang,
  and Yan Xu.
\newblock Large scale image completion via co-modulated generative adversarial
  networks.
\newblock \emph{arXiv preprint arXiv:2103.10428}, 2021.

\bibitem[Zhao et~al.(2022{\natexlab{b}})Zhao, Barnes, Zhou, Shechtman,
  Amirghodsi, and Fowlkes]{zhao2022geofill}
Yunhan Zhao, Connelly Barnes, Yuqian Zhou, Eli Shechtman, Sohrab Amirghodsi,
  and Charless Fowlkes.
\newblock Geofill: Reference-based image inpainting of scenes with complex
  geometry.
\newblock \emph{arXiv preprint arXiv:2201.08131}, 2022{\natexlab{b}}.

\bibitem[Zhou et~al.(2021)Zhou, Barnes, Shechtman, and
  Amirghodsi]{zhou2021transfill}
Yuqian Zhou, Connelly Barnes, Eli Shechtman, and Sohrab Amirghodsi.
\newblock Transfill: Reference-guided image inpainting by merging multiple
  color and spatial transformations.
\newblock In \emph{Proceedings of the IEEE/CVF conference on computer vision
  and pattern recognition}, pages 2266--2276, 2021.

\end{thebibliography}
